\documentclass[preprint,12pt]{elsarticle}

\usepackage{amssymb}
\usepackage{amsmath}
\usepackage[numbers]{natbib}
\usepackage{subfigure}

\newtheorem{problem}{Problem}
\newtheorem{scope}{Scope}

\newtheorem{definition}{Definition}

\journal{Neurocomputing}

\begin{document}

\begin{frontmatter}

\title{A${}^2$: Extracting Cyclic Switchings from DOB-nets for Rejecting Excessive Disturbances}

\author{Wenjie Lu and Dikai Liu}

\address{}

\begin{keyword}
reinforcement learning \sep
finite-state machine \sep
disturbance rejection \sep
multiple POMDPs \sep
hybrid system
\end{keyword}

\tnotetext[1]{This work was supported in part by the Australian Research Council (ARC) Linkage Project (LP150100935), the Roads and Maritime Services of NSW, and the Centre for Autonomous Systems at the University of Technology Sydney.}

%
%
%

\begin{abstract}
Reinforcement Learning (RL) is limited in practice by its gray-box nature, which is responsible for insufficient trustiness from users, unsatisfied interpretation for human intervention, inadequate analysis for future improvement, etc.
This paper seeks to partially characterize the interplay between dynamical environments and the DOB-net. The DOB-net obtained from RL solves a set of Partially Observable Markovian Decision Processes (POMDPs). The transition function of each POMDP is largely determined by the environments, which are excessive external disturbances in this research.
This paper proposes an Attention-based Abstraction (A${}^2$) approach to extract a finite-state automaton, referred to as a Key Moore Machine Network (KMMN), to capture the switching mechanisms exhibited by the DOB-net in dealing with multiple such POMDPs. This approach first quantizes the controlled platform by learning continuous-discrete interfaces. Then it extracts the KMMN by finding the key hidden states and transitions that attract sufficient attention from the DOB-net.
Within the resultant KMMN, this study found three patterns of cyclic switchings (between key hidden states), showing controls near their saturation are synchronized with unknown disturbances. Interestingly, the found switching mechanism has appeared previously in the design of hybrid control for often-saturated systems. It is further interpreted via an analogy to the discrete-event subsystem in the hybrid control.
\end{abstract}

\begin{graphicalabstract}
\includegraphics[width=1.0\hsize]{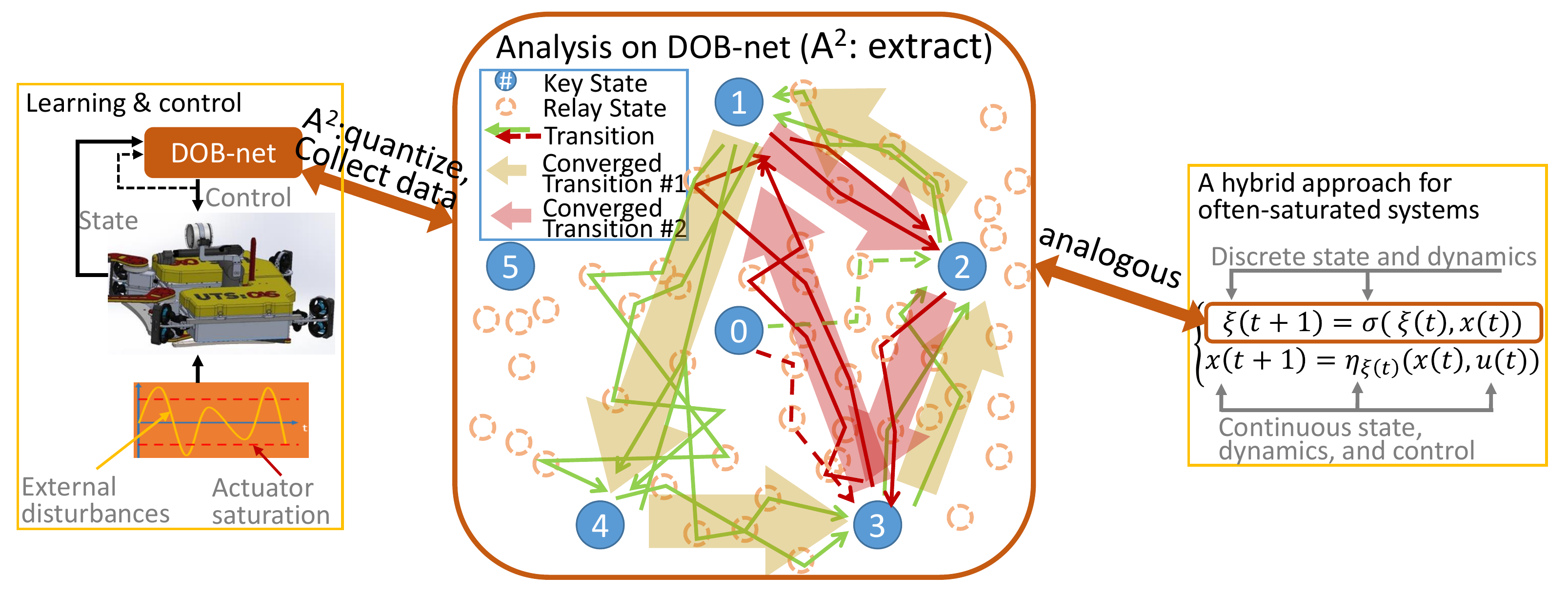}
\end{graphicalabstract}

\begin{highlights}
\item Proposed an Attention-based Abstraction (A${}^2$) approach to analyze a Disturbance OBserver network (DOB-net) that actively rejects excessive external disturbances.
\item Quantized and abstracted the learnt DOB-net via A${}^2$ and then obtained a key Moore machine network that partially reveals the interplay between the learnt control strategy and disturbances.
\item Found switching mechanisms in the resultant control for rejecting various unobservable (in statistical sense) disturbances.
\item Analyzed the captured switching mechanisms via an analogy to hybrid approaches for often-saturated systems and found that the discrete-event subsystem can be obtained by the proposed A${}^2$.
\end{highlights}

\end{frontmatter}

\section{Introduction}
\label{sec:intro}
Recent advances in deep neural networks have enabled Reinforcement Learning (RL) to solve complex problems. Model-free RL algorithms \cite{mnih2015human,oh2016control,gu2016continuous, schulman2015trust,gu2016q,lillicrap2015continuous,mnih2016asynchronous,schulman2015high} have shown their success in finding optimal control when it is difficult to precisely characterize all key elements of the targeted problems. However, the usage of RL in practical robotic applications is limited by its grey-box nature. Analytical understanding and interpretation of the learnt control networks (policies) remain unsatisfactory, particularly of those acting in continuous state, observation, and control spaces. Compared to the classical analysis of controllers and controlled systems, the learnt control networks are missing explainable control mechanisms, analytically-proved stabilities, or guaranteed asymptotical performance. In addition, there is no satisfied representation of the network internal states for fault detection and human intervention, or more importantly for knowledge distillation and transfer \cite{barreto2017successor}.

Many effort has been devoted to explainable neural networks \cite{gunning2017explainable, samek2017explainable}. Recently, finite-state representations of the learnt control networks for Atari games have been studied \cite{koul2018learning}, where each game can be viewed as a Partially Observable Markovian Decision Process (POMDP). It is found that the control policy for each game (a single POMDP) can be equivalently described by a Moore Machine Network (MMN) with states fewer than $34$, where the mapping from states to actions is described by neural networks.

While a control strategy should be able to adapt to various environments (described by a set of multiple POMDPs) online. This philosophy has found its roots in existing literature of control designs, such as robust control \cite{skogestad2007multivariable}, adaptive control \cite{aastrom2013adaptive,lu2017active,lu2018frequency}, sliding mode control \cite{edwards1998sliding}, H-infinity control \cite{doyle1989state}. Understanding the interplay between the environments and the controlled systems is essential for theoretical analysis and further improvement in control design. For example, the integration of tracking errors in Proportional-Integration-Derivation (PID) controllers and the adaptation of the compensation terms in adaptive controllers are designed to estimate some unmodelled drift online, which varies across environments. However, manual design of these estimations has limited classical controllers' applicability to complex problems, where model-free RL has shown its success.

Therefore, this paper focuses on understanding a learnt recurrent control network that is able to solve a set of multiple POMDPs, where the transition function of each POMDP is determined by an environment. The recurrent control network has to be aware of environments for effective control. In fact such a control network closely relates to the field of  ``context-aware" RL and ``meta" RL \cite{ranganathan2003middleware}. In particular, we are interested in partially understanding the dynamical interplay among environments, environment awareness, and control strategies (captured in the RL-learnt control network). This paper also seeks to find the link between the extracted interplay mechanisms and the concepts of hybrid control, which may further offer some interpretation for fault detection and human intervention.

To this end, this paper studies the position regulation problems of a free-floating underwater platform with limited thrust capacities. The platform is further subject to various excessive external disturbances. As shown in \cite{wang2019dob}, the excessive disturbance forces at consecutive two time instants can not be viewed as independently identically distributed (i.i.d) (even conditioned on the platform state). These disturbance forces are better described as unknown functions of time. Together with the platform dynamics (in static water), each disturbance defines a transition function of the disturbed platform and thus a POMDP. The disturbance and its role in determining the POMDP have to be estimated online for effective and active disturbance rejection. This paper studies a Disturbance OBserver network (DOB-net) that solves a set of such POMDPs. This DOB-net has been proposed in our previous work \cite{wang2019dob}, which integrates a disturbance observer built upon Gated Recurrent Units (GRUs). The observer subnetwork is jointly trained with a controller subnetwork for optimal disturbance encoding and rejection \cite{woolfrey2019control}.

This kind of position regulation problems arise from shallow water applications, e.g., inspecting bridge pile \cite{woolfrey2016kinematic}, and deep water operations, e.g., steering a cap to a spewing well \cite{read2011bp}, where the disturbances come from the turbulent water and oil flows may frequently exceed the thrust capability of the underwater platform. Such problems also exist in controlling quadrotors for surveillance and inspection in windy conditions \cite{waslander2009wind}. With the platform being stabilized in a range close to a targeted location, the onboard manipulators could compensate the platform oscillation and execute tasks. However, these unknown excessive disturbances inevitably bring adverse effects and may even destabilize the platform \cite{xie2000much,gao2014centrality,li2014disturbance}.

Following "less is more`` from a poem by Robert Browning, this paper proposes an Attention-based Abstraction (A${}^2$) approach for extracting key memory states and state transitions that reflect the interplay between the DOB-net and the dynamical environments (i.e., disturbances). The proposed A${}^2$ aims to equivalently present the controlled platform as a finite-state automaton. The A${}^2$ extends the Quantized Bottle Network Insertion (QBNI) \cite{koul2018learning} to the control problems that are better described by multiple POMDPs. The QBNI studies the control systems that are governed by a single POMDP with discrete observations and actions (pixels and keyboard actions). However, same to many control problems from practical applications, the position regulation problem of free-floating platforms are defined in continuous state-observation-control spaces. More critically, these problems are better described by a (possibly infinite) number of POMDPs. Therefore, A${}^2$ involves two critical improvements to the QBNI approach, as introduced below. Note that in the remainder of this paper, the terminologies ``action" and ``control" are used interchangeably.

\textbf{Contributions}: The proposed A${}^2$ first builds a discrete representation of the controlled platform by learning optimal continuous-discrete interfaces for observation and control, respectively. Since the DOB-net operates in continuous spaces, the interfaces between the discrete and the continuous spaces are required for generating KMMN. Instead of manually setting quantization levels, the A${}^2$ learns a more compact quantization that brings about minimum DOB-net performance loss. From the perspective of hybrid control, the continuous-discrete interfaces offers essential connections between continuous states and discrete modes. The switchings between discrete modes build an automaton that provides an interpretation of the switching mechanisms found later. From the perspective of machine learning, this continuous-discrete interfaces become an autoencoder with a quantization layer as the encoding layer. In this paper, these autoencoders are first trained in a supervised manner \cite{hinton2006reducing}. Then they are finetuned within the RL framework. As a result, the interfaces are optimized with the attention on the subspace where the optimally-controlled platform visits.


In addition, we propose a simple recursive loss function to train an autoencoder for quantizing hidden states, which are key in memorizing and distilling the history of the observations and controls.
We found that the autoencoder trained by the recursive loss results in a more stable DOB-net than the one trained by \cite{koul2018learning}. A Moore Machine Network (MMN) is obtained after a minimization process \cite{paull1959minimizing}, referred to as Partial Enumerative Solution (PES) for minimizing partial specified sequential switching functions. The MMN contains a large number of states and transitions, since the controlled platform inevitably undergoes multiple environments. Therefore, it is yet to have insights about the interplay between the environments and networks.

The proposed A${}^2$ further selects the MMN states and the transitions that attract sufficient attention from the DOB-net in solving multiple POMDPs, resulting in a Key MNN (KMMN). Since we are interested in the interplay between control strategies and environments (i.e., POMDPs), the attention that each state attracts is defined as the number of POMDPs that visit this state. Intuitively, a state only visited by one POMDP is unique to this POMDP and is not critical to other POMDPs. Thus this state is ignored in the KMMN. These often-visited MMN states formulate a set of KMMN states. Then the transitions between KMMN states are created if concatenated transitions between a pair of KMMN states exist. The KMMN greatly reduces the number of states and transitions, illustrating the patterns of cyclic switchings.

Within the obtained KMMN, we found that about $70\%$ of tested episodes exhibit cyclic transitions between some KMMN states. Note that each episode involves one randomly generated environment (i.e., one disturbance pattern or one POMDP). Also, we found each KMMN state corresponds to a saturated control. This finding is coherent with the fact that often-saturated systems can be described by switching-control-regulated models
\cite{zuo2010fault, benzaouia2010stabilisation, yuan2015switching, dong2010model}. We also found that the learnt control network is able to activate a portion of the KMMN, corresponding to the disturbance pattern. We still cannot fully understand the DOB-net or analyze the stability of the controlled platform. However, the induced switching mechanism may offer some interpretation of the hidden states for analysis, debugging, and abnormality detection.



In this paper, some related work is shown in Section \ref{sec:relate}. Section \ref{sec:probscope} introduces the problem formulation of the position regulation tasks, followed by the scope of this paper. Our previous work on DOB-nets is summarized in Section \ref{sec:dobnet}. Section \ref{sec:aa} provides the detailed description of the proposed A${}^2$ approach for obtaining the KMMN. Then, Section \ref{sec:results} presents the switching mechanisms found in the learnt DOB-net. The switchings are analyzed via an analogy to hybrid control in Section \ref{sec:discuss}, followed by conclusions in the last section.

\section{RELATED WORK}
\label{sec:relate}
\subsection{Disturbance Rejection}
Disturbance rejection control \cite{skogestad2007multivariable,aastrom2013adaptive,lu2018frequency,edwards1998sliding} often assumes disturbances bounded and relatively smaller than the control saturation \cite{ghafarirad2014disturbance}. One popular improvement to these controllers is to add a feedforward compensation based on some disturbance estimation techniques \cite{yang2010disturbance}. Various disturbance estimation have been proposed and practiced, such as Disturbance OBserver (DOB) \cite{ohishi1987microprocessor,chen2000nonlinear,umeno1993robust}, unknown input observer in disturbance accommodation control \cite{johnson1968optimal,johnson1971accomodation}, and extended state observer \cite{han1995extended,gao2001alternative}. However, these controllers fail to guarantee stability considering the actuator saturation \cite{gao2016nonlinear} when disturbances frequently exceed control saturation.

To this end, model predictive control (MPC) \cite{camacho2013model} is often applied due to its capability in dealing with constraints \cite{gao2016nonlinear}. It formulates a series of constrained optimization problems over receding time horizons based on predictions of the disturbed platform. A prediction method (e.g., autoregressive moving average) is required to forecast future disturbances based on the estimations of current disturbances (from DOBs). However, DOBs often require sufficient system modelling, which could be difficult for the underwater robots due to hydrodynamics effects. Otherwise, the disturbance estimations are lumped with large modelling uncertainties, which in fact are not functions of time and thus not suitable for time-series prediction methods. On the other hand, current DOBs might not have insufficient capability in estimating fast time-varying disturbances, since their convergence analysis often assume disturbances time-invariant. In addition, such separated processes of disturbance estimation, disturbance prediction, and control optimization might not be able to produce estimations and control signals that are mutual robust to each other and that jointly optimize performance, as evidenced in \cite{brahmbhatt2017deepnav,karkus2018particle}.

\subsection{Understanding Recurrent Policy Networks}
Recurrent Neural Network (RNN) memory (i.e, hidden states) is often in the form of a high-dimensional vector in a continuous space and is recursively updated through gating networks. There has been some work on visualizing and understanding the learnt RNN \cite{karpathy2015visualizing}. RNN models have been linked to iterated function systems in \cite{barnsley2014fractals}, which further shows the relationship between the independent constraints on the state dynamics and the universal clustering behavior of the network states.
Many others use training data to show the clustering and the correspondences between network internal states \cite{cleeremans1989finite}.

There has been a number of research on extracting finite-state machines from trained RNNs.
Crutchfield has reported that the minimal finite-state machine could be induced from periodic sampling with a single decision boundary \cite{crutchfield1988computation}.
An approach that forces the learning process to develop automaton representations has been proposed in \cite{frasconi1996representation}, which adds a regularization to constrain the weight space. Omlin has used hints to learn a finite-state automaton for second-order recurrent networks \cite{omlin1992training}. Learning full binary networks is an orthogonal effort \cite{hubara2016binarized} to the previously mentioned, where activation functions (and/or weights) are binary. A query-based approach has been proposed to extract a
deterministic finite-state machine that characterizes the internal dynamics of hidden states \cite{weiss2017extracting}.

Koul has proposed Quantized Bottleneck Network (QBN) insertion in \cite{koul2018learning} for extracting a finite-state machine from discrete action networks. The QBNs are autoencoders, where the latent encoding is quantized. Given a trained RNN policy, the QBNs are trained to encode and quantize the hidden states and observations in a supervised manner.

\section{CONTROL PROBLEM and SCOPE}
\label{sec:probscope}
The optimal control problems considered in this study involve a free-floating platform (a rigid body) under translational motion, thanks to its huge restoring forces and its sufficiently large torque capacity on the heading control. The position of this platform is denoted as $q \in \mathbb{R}^3$. The platform's velocities and accelerations are denoted by $\dot{q}$ and $\ddot{q} \in \mathbb{R}^{3}$, respectively. It is assumed that $q$ and $\dot{q}$ are observable without errors in this study, nevertheless RL approaches are in general robust to reasonable observation noises.

Then the platform's dynamics (also referred to as the system) is given by
\begin{equation}
M(q) \ddot{q}+g(q)=u+d,
\end{equation}
where $M(q) \in \mathbb{R}^{3 \times 3}$ is the inertia matrix and $g(q) \in \mathbb{R}^{3}$ is the vector of the gravity and buoyancy forces. This matrix , vector, and external disturbances $d \subset\mathbb{R}^{3}$ are assumed unknown to the controller (the trained DOB-net). The platform control $u \in \mathcal{A}\subset\mathbb{R}^{3}$ is saturated at an upper bound $u^{-} =\max (\mathcal{A})\in \mathbb{R}^{3}$ and a lower bound $u_{-} = \min (\mathcal{A}) \in \mathbb{R}^{3}$, where $\max$ and $\min$ are dimension-wise operators. Let $x=[q^T,\dot{q}^T]^T\in\mathcal{X}\subset\mathbb{R}^6$ and the platform dynamics in discrete time can be written as
\begin{equation}
\label{eq:difference}
x(t+1) = f\big(x(t),u(t),d(t)\big).
\end{equation}
In the remainder of this paper, the time indices $t$ in equations are in parentheses and the ones in figures are subscripts for compactness.

The external disturbances $d$ are described by the disturbance forces, which are time-variant and are superpositions of $l$ sinusoidal functions as
\begin{align}
\label{eq:disturbance}
d(t) = \sum_{1 \le i \le l} d_i(t),
\end{align}
where $d_i(t) = A_i \sin (w_i t+\phi_i)$, and $l$ denotes the number of components, which is unknown and may vary across environments. The parameters ($A_i$, $w_i$, and $\phi_i$) of each component $d_i$ are assumed uniformly and randomly sampled from given intervals and then fixed in each environment in this paper. One instantiation of all parameters of all $l$ components is referred to one {\it disturbance pattern}.

In the remainder of this paper, one sampled disturbance pattern is viewed as one environment to the free-floating body. The terms ``disturbances" and ``disturbance forces" are used exchangeably. Note that the external disturbances considered are excessive to the free-floating platform, the definition of which is given as follows.

\begin{definition}[Excessive external disturbances]
Excessive external disturbances are those defined in eq. (\ref{eq:disturbance}), where the amplitudes ($A_i$) exceed the platform control saturation ($u^{-}$ and $u_{-}$).
\end{definition}

\begin{problem}[Optimal control]
\label{pb:oc}
Find one controller that chooses an action $u(t)$ for the system described in eq. (\ref{eq:difference}) at time $t$ in response to the current observation $x(t)$, such that the discounted summation of collected rewards is maximized. The summation is expected over episodes and is defined as $\mathbb{E}\sum_{\tau=t}^{T-1} \gamma^{\tau-t} r\big(x(\tau),u(\tau)\big)$, where $r(\cdot)$ is a reward function (additive inverse of the tracking error, defined later), $T$ denotes number of steps in an episode, and $\gamma \in [0, 1)$ is a discount factor that prioritizes near-term rewards over future rewards \cite{nagabandi2018neural}.
\end{problem}

The tracking error is defined as
\begin{align}
\eta(t) = \|x(t)\|,
\end{align}
where $\|\cdot\|$ denotes the L$2$ norm. In each episode, the environment is randomly sampled and is characterized by excessive disturbances in eq. (\ref{eq:disturbance}).

Classical RL approaches often implicitly assume $d(t)$ independently identically distributed (i.i.d.), possibly conditioned on the platform state $x$ \cite{saemundsson2018meta}. If not conditioned on $x$, $d(t)$ is marginalized over $t$ and $x$, and is then described as $d(t) \sim \mathcal{N}(\epsilon, E)$, where $E = \operatorname{diag}\left(\sigma_{1}^{2}, \ldots, \sigma_{D}^{2}\right)$. These disturbance models lead to a single-POMDP description of the controlled platform and is sufficient when disturbances are small. However, the excessiveness makes these models of $d$ not suitable for disturbance rejection, as evidenced in \cite{wang2019dob}. The following analysis shows that the controlled systems in Problem \ref{pb:oc} are better described by multiple POMDPs.

For a $j$th pattern of disturbance superposition, each component $d_{i,j}(t)$ of $d_j(t)$ is a function that exhibits periodicity, which can be described as a Markovian chain. The index $j$ is dropped if no ambiguity is caused. The Markovian chain is given as
\begin{equation}
\begin{bmatrix}d_i(t+1)\\\dot{d}_i(t+1)\end{bmatrix} = g_{i,j}\big(\begin{bmatrix}d_i(t)\\\dot{d}_i(t)\end{bmatrix}\big), ~1 \le i \le l,
\end{equation}
where the index $j\in\{1,2,\cdots,\infty\}$ of $g_{i,j}$ indicates the variety of disturbance patterns. Let $\mathcal{D}\subset\mathbb{R}^6$ denote the space of $[d^T_i(t+1), \dot{d}^T_i(t+1)]^T$ and $\mathcal{G}$ the space of possible $g_{i,j}$.

Let $z_j=[x^T,~d^T_{1,j},~\dot{d}^T_{1,j},~\cdots,d^T_{l,j},~\dot{d}^T_{l,j}]^T\in\mathcal{Z}_j=\mathcal{X}\times\mathcal{D}^l$,
where $l$ might vary across environments. Then the platform dynamics can be rewritten in a partially observable Markovian chain as
\begin{align}
\label{eq:problem}
z_j(t+1) &= f_j\big(z_j(t),u(t)\big),\nonumber\\
y(t) &= h_j\big(z(t)\big) = x(t),
\end{align}
where $h_j(\cdot)$ is the observation function, showing that $x(t)$ is observable while $d(t)$ is not directly observable. Here the observability is in statistical sense (not in control sense). Let $\mathcal{F}$ denote the space of all possible $f_j$. Each transition function $f_j$ defines a POMDP $P_j = \{\mathcal{Z}_j,\mathcal{A},f_j,h_j,\pi\}$, where $\pi$ is the trained current control network. Let $\mathcal{P}$ denote the set of all possible $P_j$.

The control network $\pi$ is targeted to solve Problem \ref{pb:oc} (i.e., all $P_j\in\mathcal{P}$). Key to $\pi$ is the integration of a disturbance observer to existing RL frameworks. This observer not just estimates the unobservable state $d_j(t)$ but also infer the transition function $f_j$. Both $d_j(t)$ and $f_j$ are critical to the control subnetwork. Our previous work has proposed a DOB-net for this purpose \cite{wang2019dob}. The DOB-net outperforms existing control and RL approaches. However, the understanding of the learnt DOB-net remains unsatisfactory. Therefore, the scope of this paper, shown below, is regarding the understanding of the learnt DOB-net. For simplicity, the reduced version of the DOB-net is studied in this paper.

\begin{scope}[Analysis of DOB-nets]
\label{def:scope}
Inductive reasoning of the mechanism on how the learnt DOB-net responds to different unobservable external excessive disturbances (i.e., to different POMDPs).
\end{scope}

\section{DOB-NET}
\label{sec:dobnet}
\begin{figure}
	\centering
	\includegraphics[height=4.8cm]{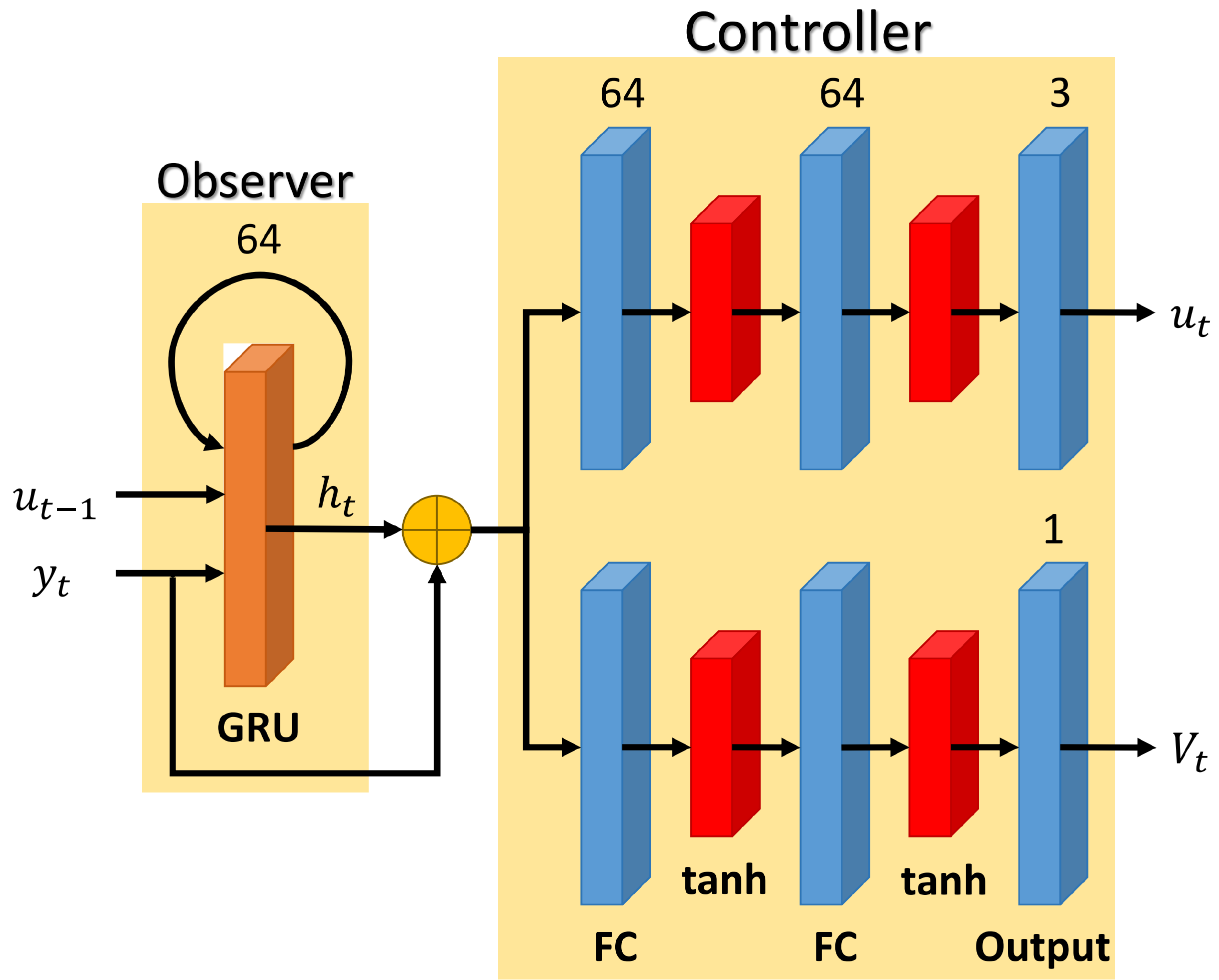}
	\caption{Network architecture of DOB-Nets. FC: fully connected layer.}
	\label{fig:dob_net}
\end{figure}

Estimating the disturbance forces, their transition functions, and their predictions is key in solving a $P_j$ randomly sampled from $\mathcal{P}$. In DOB-nets, these estimations are encoded in a latent feature space. The features have to be mutual robust between the controller and the observer. The DOB-net developed in our previous work is composed of a disturbance-behaviour observer subnetwork and a controller subnetwork. For simplicity, this paper investigates the reduced version consisting of a single-layer GRU, as shown in Fig. \ref{fig:dob_net}. Both subnetworks are jointly optimized for mutual robustness and unified optimization.
The observer subnetwork imitates the classical DOB mechanisms and is enhanced with the flexibility from GRUs, instead of only providing the estimation of the lumped disturbances up to the current time. The encoding $h_{t}$ (shown in Fig. \ref{fig:dob_net}) is supposed to represent the disturbance behaviour that is key to controller subnetwork.

The full DOB-net is constructed based on the classical actor-critic architecture \cite{LuADPSwitched14}, the network outputs actions and critics (also referred to as cost-to-go) associate with previous state and action. The policy is trained using simulated sine-wave disturbances. Multiple control and RL algorithms have been tested and compared in \cite{wang2019dob}, the results have demonstrated that the proposed DOB-net does have a significant improvement in rejecting excessive disturbances.

\section{A${}^2$: Extracting Key Moore Machine Network}
\label{sec:aa}
The proposed A${}^2$ approach aims to abstract the control mechanism captured in the trained DOB-net for solving continuous-control problems. It consists of two procedures: quantization and abstraction. The A${}^2$ involves two critical improvements to the ``Quantized Bottleneck Network Insertion" (QBNI) \cite{koul2018learning}, the later of which is used to generate a finite-state automaton of a trained policy network.

\begin{definition}[Finite-state automaton \cite{cheng1993automatic}]
A finite-state automaton is an abstract machine whose state is assigned as one of a finite number of states at any given time. It is also referred to as a finite-state machine. The transition between states is determined by discrete action and observation, which is often given by a table.
\end{definition}
The finite-state automata in this paper are all deterministic.

\begin{definition}[Moore machine network \cite{koul2018learning}]
A Moore machine network is a standard deterministic finite-state machine whose states are labeled by their output values (controls in this paper). A MMN is fully characterized by finite sets of states, observations, and actions, a transition function, and a policy that maps states to actions, where the policy and the transition function are represented by neural networks.
\end{definition}

The QBNI algorithm together with the PES work well for grouping hidden states and observations (and thus reducing number of states in a MMN). However, the effectiveness of the PES heavily depends on the number of actions, which has to be limited. At least one state is related to a unique action \cite{paull1959minimizing}, therefore the number of possible actions have to be reduced for revealing the interplay in Scope \ref{def:scope}. In the cases of Atari games, the possible actions are often fewer than $8$ (e.g., ``fire", ``move left/right", ``jump"). As pointed in the introduction, the problems studied here involve multiple POMDPs, leading to a large number of states and transition in the obtained MMN.

The first improvement is in the quantization, where continuous-discrete interfaces are optimized for actions, reducing the number of quantized actions given acceptable DOB-net performance loss. The second improvement is the abstraction of key states and transitions in the MMN based on the evaluation of attention.



\subsection{Continuous-Discrete Interfaces}
The proposed A${}^2$ approach first learns continuous-discrete interfaces for observations and action, respectively. These interfaces commonly exist in hybrid system modelling \cite{lu2015hybrid} for solving control of often-saturated systems. The observation and action interfaces in the quantized DOB-net have been shown in Fig. \ref{fig:interfaces} (better viewed in color), which are denoted as Observation Quantization (OQ) and the Action Quantization (AQ), respectively. Each Quantization block consists of a continuous-to-discrete interface and a discrete-to-continuous interface.

Then, the components in the blue dashed rectangle and the ones in the green dash-dotted rectangle correspond to the discrete-event subsystem and the mapping from the discrete hidden state to the continuous control, respectively. This will be discussed more in Section \ref{sec:results}. In this paper, all interfaces are built upon neural networks, the detailed structures of which are shown in Section \ref{sec:results}. In fact each quantization block is an autoencoder from the perspective of machine learning.

In general autoencoders consist of an encoder and an decoder, where the decoder aims to reconstruct the original inputs to the encoder. The autoencoder has been used widely to reduce data dimension using neural networks \cite{hinton2006reducing}, which is often trained in a supervised manner.

One straightforward approach to have the interfaces is to evenly quantize the observation and action space, however, the quantization levels are not clear. Also, the importance of action (observation) to the controlled platform is not uniform across the action (observation) space. The states and actions that attract most attention from the optimally-controlled platform are often subsets of the entire state and action spaces, respectively. We are interested in the interfaces that are both optimized with respective to these subsets.

In this paper, to produce a continuous-to-discrete interface, the output of the encoder is quantized through a combination of a $3$-level activation layer (denoted as Tanh*) and a quantization layer. Same to \cite{koul2018learning}, the Tanh* layer restricts the outputs in the range of $[-1,1]$ and offers $0$ gradient near a $0$-valued input, which allows a quantization level at $0$ during training. The Tanh* activation function is given as \cite{koul2018learning}
\begin{align}
\label{eq:tanhs}
\phi(x) = 1.5 \mbox{tanh}(x)+0.5 \mbox{tanh}(-3x).
\end{align}
With Tanh*, the quantization layer offers $3$-level quantization valued at $\{+1,0,1\}$.

\begin{figure}
	\centering
	\includegraphics[height=4.8cm]{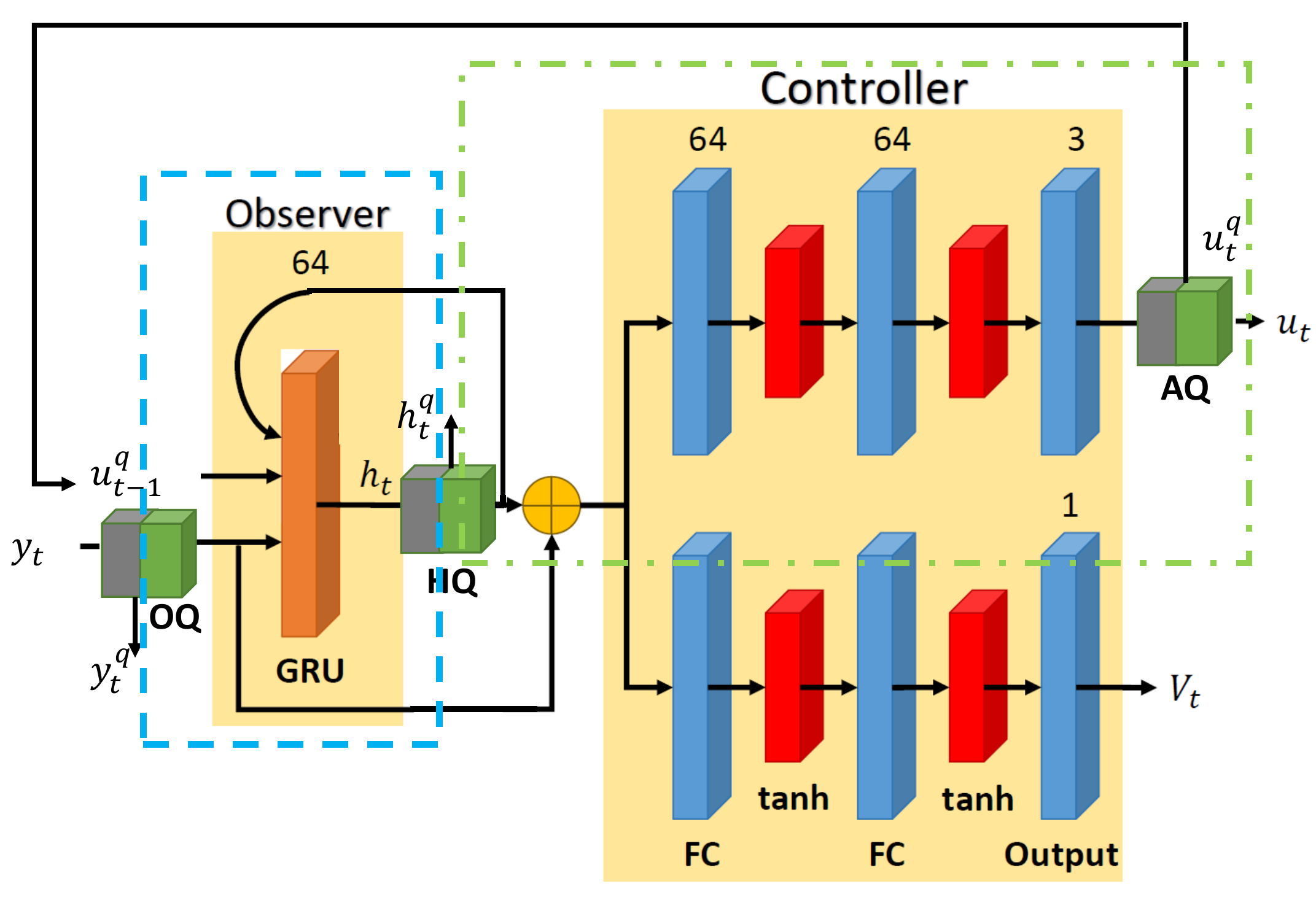}
	\caption{Network architecture of the quantized DOB-Net.}
	\label{fig:dob_quantized}
\end{figure}

With the continuous-discrete interfaces inserted, the full quantized DOB-net is illustrated in Fig. \ref{fig:dob_quantized}. In the remainder of this paper, the original DOB-net is referred to as ``continuous DOB-net" to distinguish from the quantized DOB-net.


\textbf{Training}: The QBNI algorithm, suggested in \cite{koul2018learning}, does not work well for learning OQ and AQ, since the number of quantized actions should also be minimized for effective reduction in obtaining a key MMN.
Therefore, a three-step training approach is used to train both OQ and AQ. The number $e$ of neurons in the encoder layer of AQ determines the cardinality of the set of all possible discrete actions. The cardinality is $3^e$ since each quantization neuron has $3$ levels. On one hand, a large $e$ leads to less optimality loss from the quantization, compared with the continuous DOB-net. On other hand, a small $e$ results in fewer action choices and thus fewer MMN states after reduction by PES. Therefore, the number of discrete actions is expected to be minimized for the benefit of reducing the number of states in the MMN. By choosing the number of neurons in the quantization layer, the performance degeneration should be restricted within a reasonable number (e.g., $10\%$).

{\it Step one}: The continuous DOB-net is first trained by the Advantage Actor Critic (A2C) \cite{mnih2016asynchronous}, as shown in \cite{wang2019dob}. A2C uses synchronous gradient descents for optimizing policy networks and it executes multiple instances of the environments in parallel threads. This parallelism provides a more training estimation of critics.

{\it Step Two}: A data set of observations and actions from a large number of episodes is collected through using the trained continuous DOB-net. Note that in each episode, a disturbance pattern is randomly generated, which is i.i.d. to the pattern in another episode. Then OQ and AQ are trained respectively using the observation and action data through supervised learning. Since the data is collected from using the optimal DOB-net, the data reflects the nonuniform distribution of attention in the action and observation space.

{\it Step Three}: The trained OQ and AQ are inserted into the trained continuous DOB-net to obtain the quantized DOB-net, as shown in the Fig. \ref{fig:dob_quantized} (HQ is deactivated). However, the performance of the quantized DOB-net is not close to the continuous DOB-net (worse by $31\%$). Then the entire quantized DOB-net is finetuned in a RL fashion, same to Step One. The quantization layer introduces functions that are non-differentiable. During the training, a straight-through estimator for gradients, as suggested in \cite{bengio2013estimating}, is adopted. The estimator simply treats the quantize function as an identity function during back propagation and passes on the gradients without any change.
The results shown in Section \ref{sec:results} suggest that the performance of the quantized DOB-net resulted from the three-step training is close to the performance from the continuous DOB-net.

Once the quantized DOB-net (HQ deactivated) is obtained, a data set of hidden states is collected and used to train HQ, as in \cite{koul2018learning}. With the trained HQ insertion as illustrated in Fig. \ref{fig:dob_quantized}, the full quantized DOB-net is available.



\subsection{Key Moore Machine Network}
\label{sec:kmmn}
The data sets of the discrete hidden states, the discrete observations, and the discrete actions are collected during solving Problem \ref{pb:oc} in multiple randomly-generated environments. In addition, the transitions between consecutive pairs of the quantized hidden states are also recorded.

Then unique states are found and indexed for each data set, resulting in a MMN.
Let $m$ denote the cardinality of the state space of the MMN and $n$ the cardinality of the observation space of the MMN, then the transition function of this MMN is constructed as a transition matrix of $n\times m$ that captures the transitions evidenced in the data. In general, $m$ and $n$ are larger than necessary.

A reduced but equivalent MMN can be obtained by a standard finite state machine reduction technique (i.e., PES in this paper), which is able to group hidden states and observations if a common transition and action can be found. Each group of the hidden states is referred to as a state in the reduced MMN and each group of the observations is referred to as an observation in the reduced MMN. This reduced MMN is able to show how states, observations, and actions are related to problems, as shown in \cite{koul2018learning}.

\begin{figure}
	\centering
	\includegraphics[width=0.5\hsize]{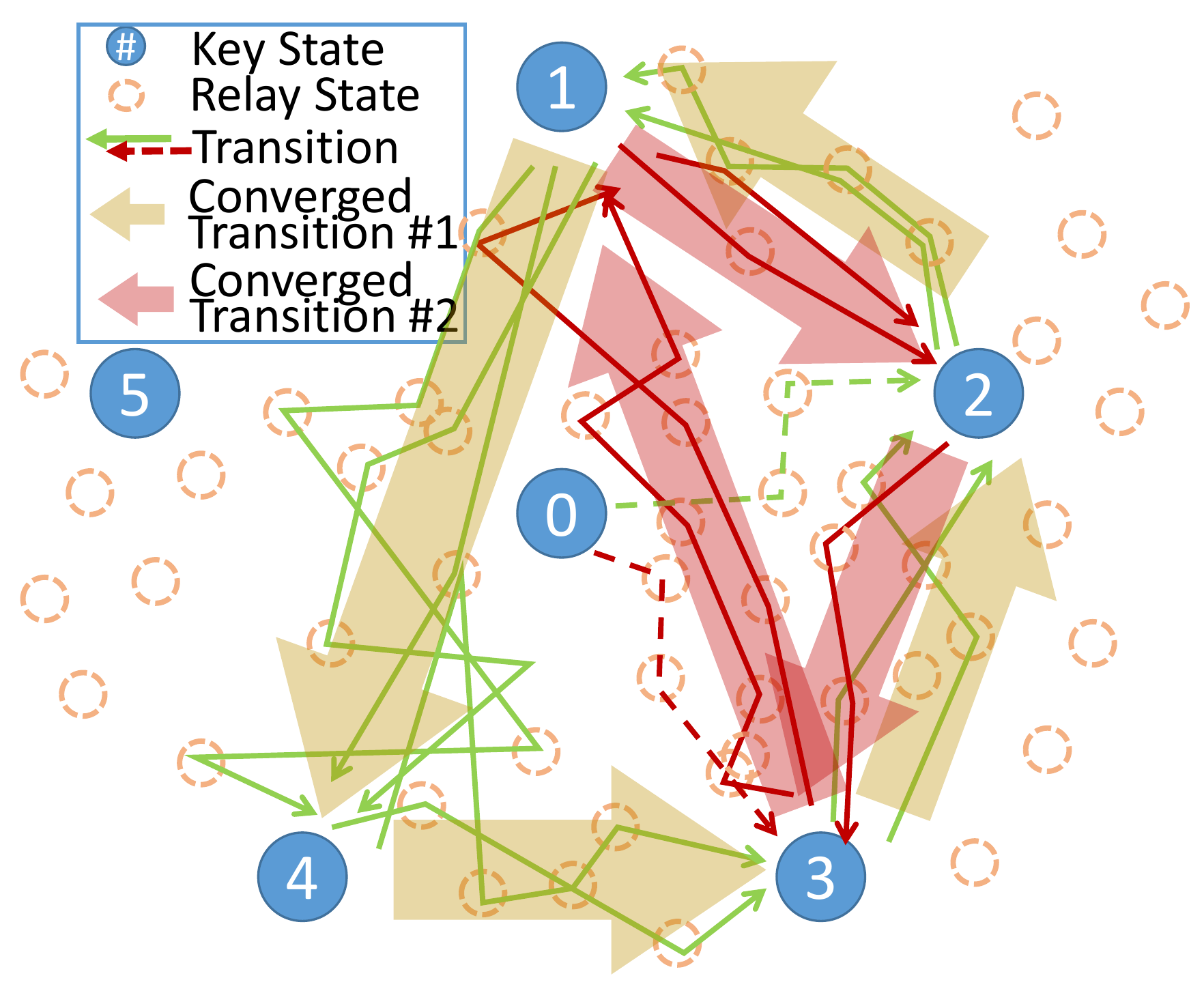}
	\caption{Key transitions: part of the reduced MMN.}
	\label{fig:kmmn}
\end{figure}

However, Problem $1$ subject to various environments are better described by multiple randomly sampled POMDPs. The number of states and observations in the reduced MMN are still too large to induce explainable relationship among states, action, and environments. In fact, the systems (controlled by the quantized DOB-net) visit different portions of the reduced MMN in different episodes (i.e., under various disturbance patterns), as illustrated in Fig. \ref{fig:kmmn}. As shown in Section \ref{sec:results}, the number of states in the reduced MMN was still quite large ($91$) compared to Atari games investigated in \cite{koul2018learning}.

In order to understand the interplay between disturbances and control strategies, in this paper, we propose a Key Moore Machine Network (KMMN), which ignores some states and transitions in the reduced MMN. Some of the states and observations are unique to an episode (i.e, a POMDP), while others attract more attention from a number of episodes.

\begin{definition}[Key Moore machine network]
A key Moore machine network is a finite-state automaton that only consists of the key states and transitions between key states. The key state are those MMN states that attract sufficient attention from the controlled systems in a number of environments.
The attention of a state is defined as the number of episodes that visit this state. A transition between the key states is available if a concatenated transition can be found in the reduced MMN.
\end{definition}

The relation between the KMMN and the reduced MMN is shown in Fig. \ref{fig:kmmn}, where the MMN states other than key states are referred to as relay states.
Note that the transitions in KMMN are different from the MMN transitions. A KMMN transition may involve multiple MMN transitions. One MMN transition corresponds to one step defined in POMDPs. Since we are interested in the interplay between the control strategies and the environments (i.e., disturbances, POMDPs), we extract key MMN states that are commonly visited by a number of POMDPs. The KMMN greatly reduces the number of states and transitions, providing a baseline for inductive learning of the interplay. To find KMMN, the step of obtaining the reduced MMN is necessary. Otherwise the chance of having states with sufficient attention is quite low.

\section{IMPLEMENTATION and RESULTS}
\label{sec:results}
This section first outlines the simulation details of the platform and disturbances. Then, the implementation of learning interfaces (AQ, OQ, and HQ) and results are presented. After that the obtained MMN and KMMN are summarized, as well as the found switching mechanism captured in the DOB-net.

\subsection{Platform and Disturbances}
As described in the problem formulation, the platforms is assumed stable in orientation. Only translational motion and control are considered, thus, the platform has a $6$-dimensional state space (positions and linear velocities) and a $3$-dimensional action space. In order to analyze the results more intuitively, the characteristics (mass, control, gravity and buoyancy forces, and disturbance forces) of the platform are scaled down such that the mass of the simulated platform is $1$ [kg].  Then, the control saturation is given as $u^{-} = -u_{-} = [2,~ 2,~ 2]^{T} ~[N]$.

Each episode contains $200$ steps with $0.05$ second per step. In each episode, the platform starts at a random position with a random velocity, and it is controlled to reach a given position (the origin), aiming to keep its position within a range (as small as possible) to the origin against unknown excessive disturbances. In these simulations, the external disturbances are exerted on all three directions in the inertial frame. In each axis, the disturbance is sinusoidal and then the disturbance superposition is given as
\begin{align}
d(t) = \begin{bmatrix} A_x\sin(\frac{\pi}{T_x} t+\phi_x)\\ A_y\sin(\frac{\pi}{T_y} t+\phi_y)\\A_z\sin(\frac{\pi}{T_z} t+\phi_z)\end{bmatrix},
\end{align}
where
\begin{align}
\label{eq:rand_dist}
A_x, A_y, A_z &\sim U(2.6,3)\nonumber\\
T_x, T_y, T_z &\sim U(2,4)\nonumber\\
\phi_x, \phi_y, \phi_z &\sim U(-\pi,\pi),
\end{align}
and $U(a,b)$ denotes a uniform distribution in the range $[a,~b]$. According to the problem setting, the amplitudes of disturbances exceed the control limits by $30\%-50\%$. The purpose of the DOB-net training is to enable the trained network to deal with unknown time-varying disturbances, thus the values of the amplitude, period, and phase are randomly sampled in each training or testing episode.

\subsection{Learning Interfaces}
\begin{figure}
	\centering \includegraphics[width=0.7\hsize]{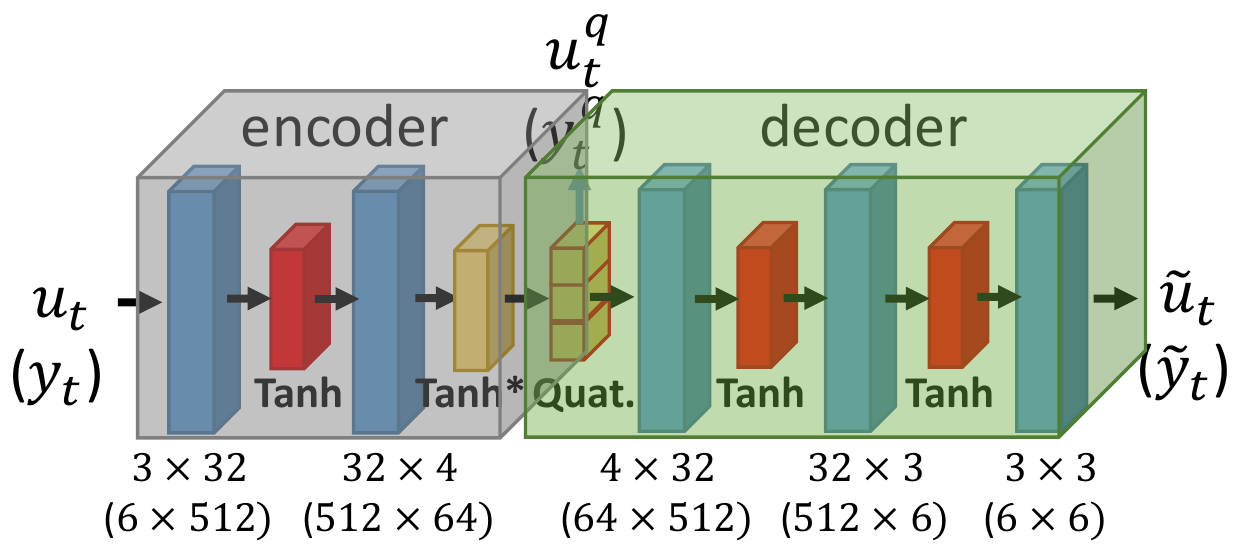}
	\caption{Network structure of AQ and OQ. Quantities in parentheses correspond to OQ.}
	\label{fig:interfaces}
\end{figure}

The interfaces for action (AQ) are illustrated in Fig. \ref{fig:interfaces}, which consists of $5$ linear layers, $1$ quantization layer, and $4$ hyperbolic tangent (denoted as Tanh) activation layers. One of the activation layers is a $3$-level activation layer (defined in eq. (\ref{eq:tanhs}) and denoted as Tanh*). The encoder component of the autoencoder is a continuous-to-discrete interface, while the decoder component is a discrete-to-continuous interface. The interfaces for action and observation share a similar autoencoder structure with different numbers of neurons in linear layers and the quantization layer. In Fig. \ref{fig:interfaces}, the numbers and symbols in parentheses show the input, the output, and the number of neurons regarding OQ.

The neuron numbers were manually picked such that the quantized DOB-net performs similarly to its continuous counterpart. As pointed out earlier, the number of neurons in the encoding layer of AQ is critical. It is expected to minimize this number without loosing much optimality in the resultant quantized DOB-net. It was manually picked via trail-and-error approach. The neuron number was first set to $3$, however the resultant performance was not satisfactory. The collected reward (negative) was nearly doubled. Then, the neuron number was set to $4$ and $5$, respectively. It was found that $4$ is sufficient for retaining optimality. The continuous DOB-net and quantized DOB-net exhibit on average $10\%$ difference in rewards collected in an episode. The number of neurons in OQ is also critical, choices of $32$, $48$, $64$, and $128$ were tested and it was found that $64$ is an appropriate for the DOB-net. The choice of neuron numbers has been studied in the field of neural architecture search and can possibly be solved via RL \cite{zoph2016neural}, however it is out of the paper scope.

\begin{figure}
	\centering \includegraphics[width=0.5\hsize]{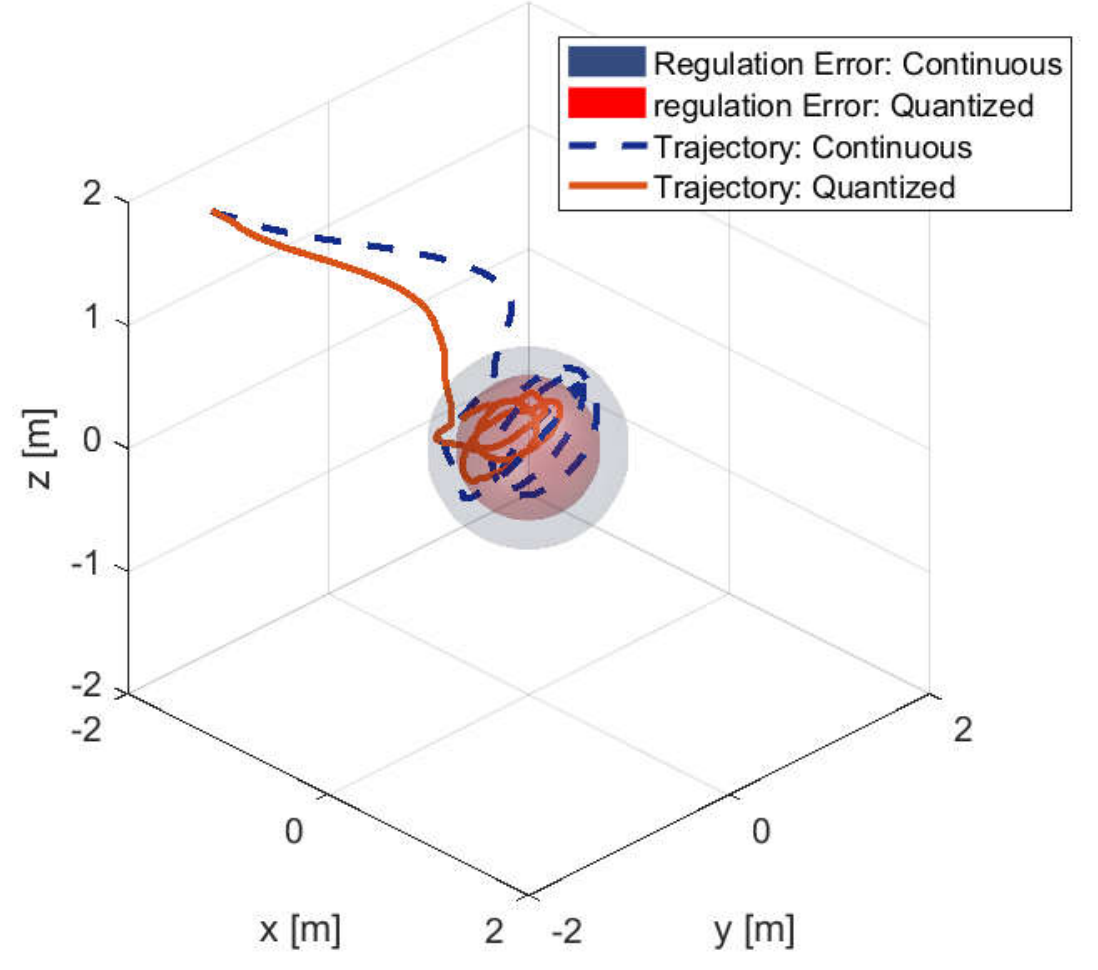}	\caption{Comparison of example trajectories from the continuous DOB-net and the quantized DOB-net.}
	\label{fig:traj1}
\end{figure}
\begin{figure}
	\centering \includegraphics[width=0.5\hsize]{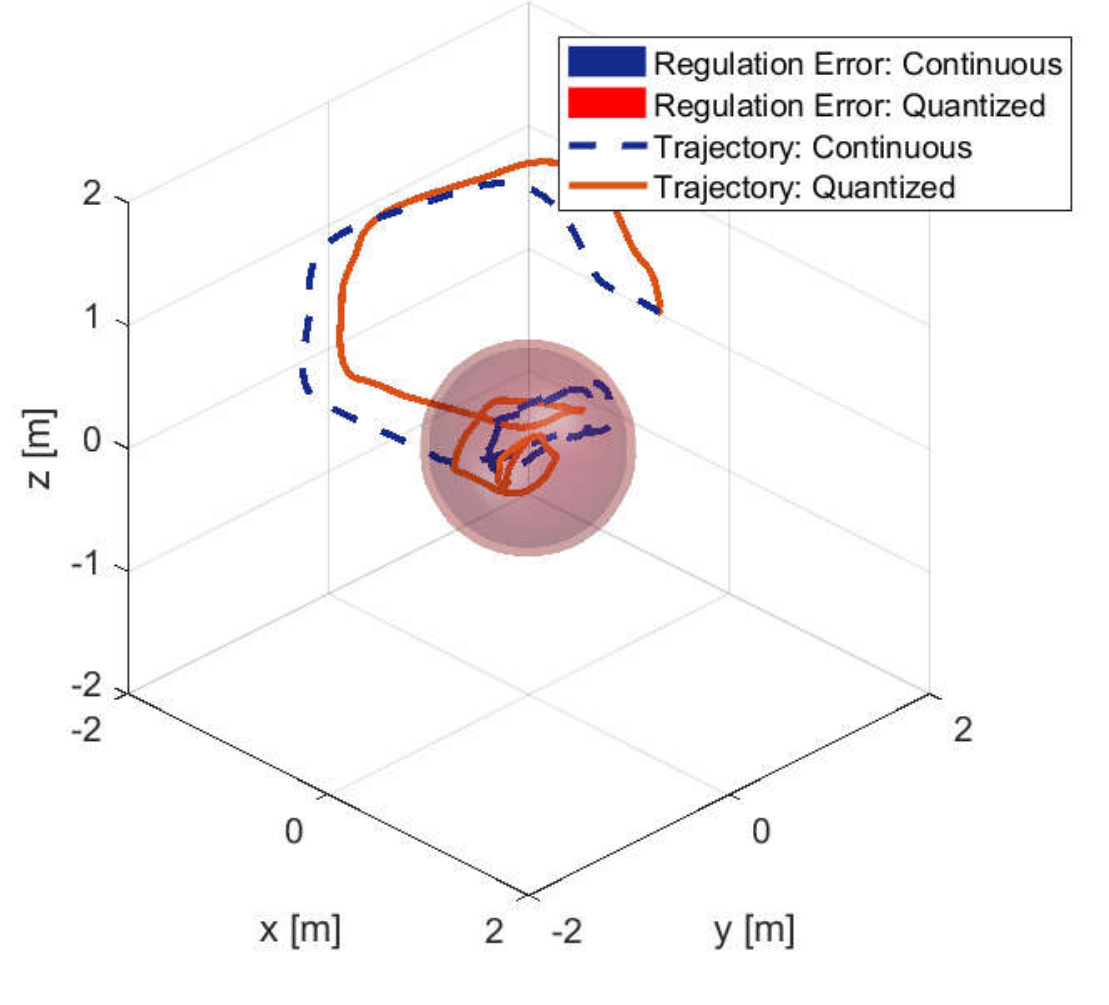} \caption{Comparison of example trajectories from the continuous DOB-net and the quantized DOB-net.}
	\label{fig:traj4}
\end{figure}

Since the disturbances exceed the control saturation frequently, the platform inevitably oscillates and so does the error of position regulation. The DOB-net requires some steps to collect sufficient data to infer the environment in the hidden state. Here the maximum tracking error from $t=150$ to $t=200$ is used as one criteria to show the effectiveness of the learnt DOB-nets. It is referred to as the regulation error and given as
\begin{align}
\label{eq:error}
R = \max_{t} \eta(t), 150 \le t \le 200.
\end{align}
The $3$D trajectories from both quantized and continuous DOB-nets for same problems (i.e., same POMDPs defined in eq. (\ref{eq:problem})) have been illustrated in Figures~\ref{fig:traj1} and \ref{fig:traj4}, respectively. The transparent red and blue spheres respectively represent the regulation errors from the quantized and continuous DOB-nets. Clearly, the quantized DOB-net was able to achieve trajectories similar to the one from the continuous DOB-net. Furthermore, the regulation error did not increase much.

\subsection{Moore Machine Networks}
\begin{figure}
	\centering \includegraphics[width=0.7\hsize]{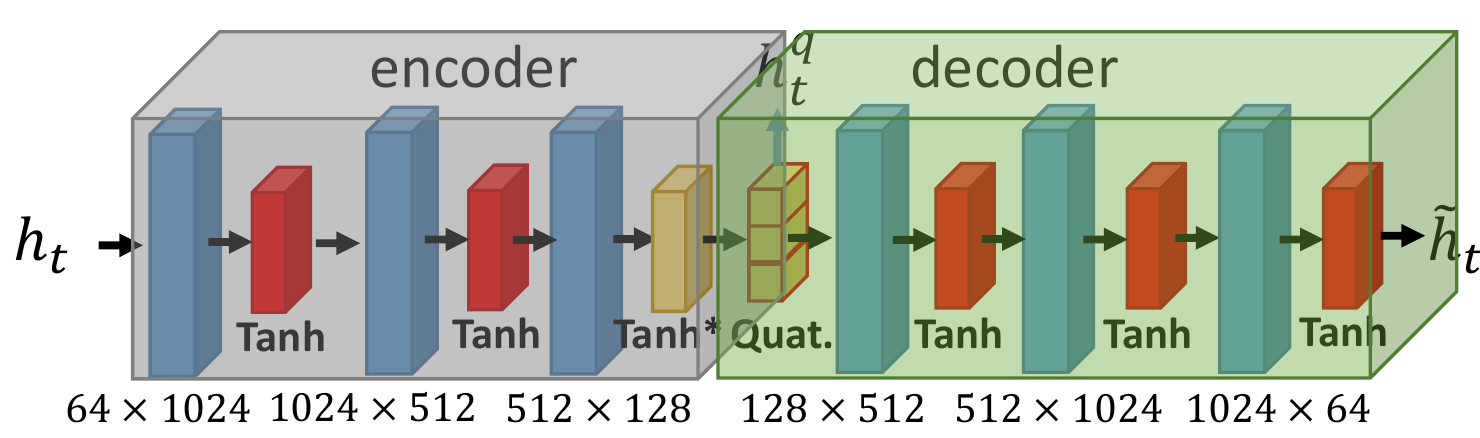}
	\caption{Hidden state quantization.}
	\label{fig:hx}
\end{figure}

Once the interfaces for action and observations was trained, another set of simulations using the quantized DOB-net were conducted. A data set of the GRU hidden states was collected from $1000$ episodes. In each episode, the disturbance pattern was randomly generated according to eq. (\ref{eq:rand_dist}). Following \cite{koul2018learning}, the autoencoder for quantizing hidden states is illustrated in Fig. \ref{fig:hx}, which consists of $6$ linear layers, $1$ quantization layer, and $6$ Tanh activation layers, where one of the activation layer is Tanh*.

\begin{figure}
	\centering \includegraphics[width=0.7\hsize]{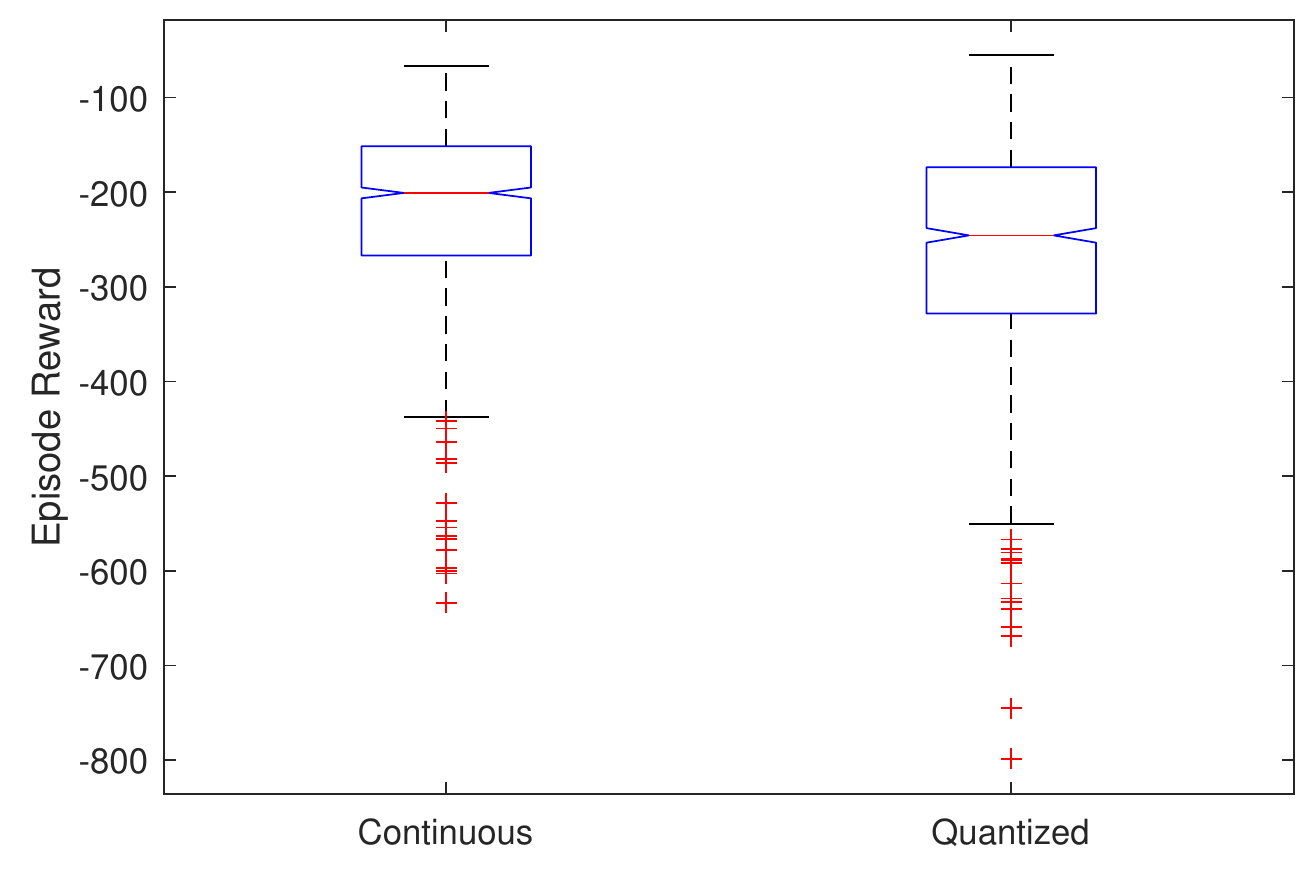}
	\caption{Rewards collected from $1000$ episodes.}
	\label{fig:test_reward}
\end{figure}

\begin{figure}
	\centering \includegraphics[width=0.7\hsize]{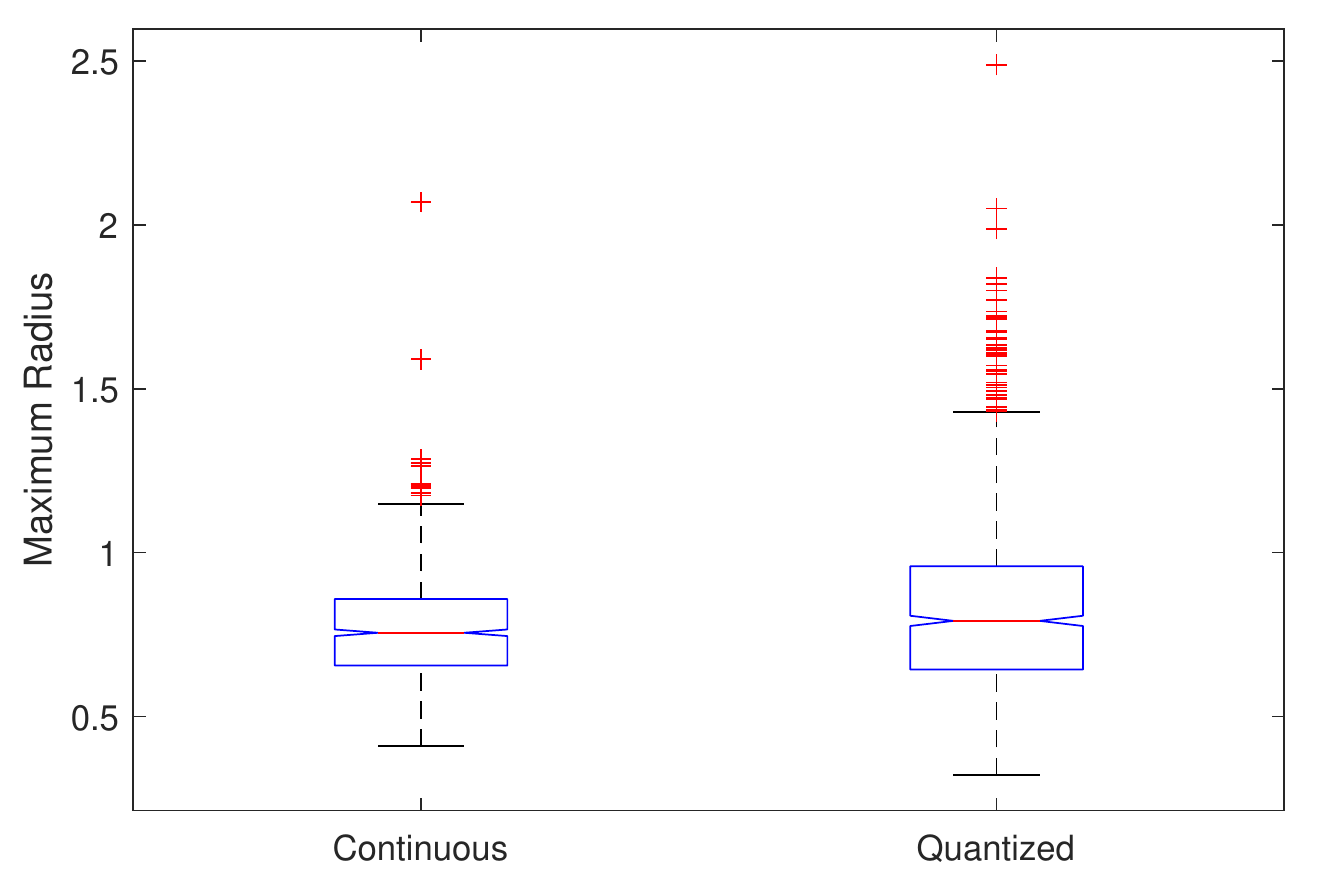}
	\caption{Position regulation errors collected from $1000$ episodes.}
	\label{fig:test_distance}
\end{figure}

The data collected was used to train HQ in a supervised manner. Different from usual loss functions, the importance of recursive stability was emphasized. The loss function used has two terms; the first one is standard and the second one regulates the recursive stability. The loss function $L$ is defined as
\begin{align}
L = ||h_t - HQ(h_t)||+ \eta||h_t - HQ(HQ(h_t))||,
\end{align}
where $\eta$ was set as $10$. Using stochastic gradient descent approach with the learning rate $1e\!-\!4$, the training error (mean square error) was $1.2e\!-\!3$. The HQ network was inserted into the quantized DOB-net, as suggested in \cite{koul2018learning}, resulting in the full quantized DOB-net. The rewards collected in each episode by the quantized DOB-net has been compared with the ones collected by the continuous DOB-net in Fig. \ref{fig:test_reward}, showing about $12\%$ degeneration averaged over all episodes. As shown in Fig. \ref{fig:test_distance}, the averaged regulation error exhibited $8\%$ increase.

Then another data was collected from simulations of $400$ episodes using the quantized DOB-net (with HQ inserted). Each episode has $200$ samples of observations, hidden states, current actions, and previous actions. Also the transitions between hidden states given observations and actions were recorded. It was found that the number of the unique hidden states was $16931$ and the number of the unique observations was $15619$ , suggesting that the system controlled by the quantized DOB-net in multiple environments did visit a large number of discrete hidden states. The number of the unique actions was $80$, the maximum of which is $81$.

Considering the transitions between discrete hidden states as an incompletely specified sequential switching functions, the number of hidden states and observations was grouped by PES \cite{paull1959minimizing}. The number of unique groups of hidden states in the reduced MMN was reduced to $114$ and the number of observation groups was $2047$.  We refer to each of this group as a state or a observation in MMN. It is nearly impossible to find insights about the interplay between environments and control strategies, due to the large number of the transitions and states. A portion of the MMN that highlights the transitions and states visited by two episodes has been illustrated in Fig. \ref{fig:kmmn}, where the key states were obtained in the following subsection.



\subsection{Key Moore Machine Network}
The goal of the KMMN is to extract some shared control logics used by the learnt DOB-net to solve different POMDPs defined in eq. (\ref{eq:problem}), and thus to show the interplay between the control and disturbances. Here data from $20$ episodes were studied. The sufficient attention was defined as ``$85\%$ attention". In other words, being qualified as a key state in KMMN, the state must attract attention from at least $17$ episodes out of $20$.

We found that $6$ key states were picked by those $20$ episodes. One of the key states is the initial state since in all episodes the hidden state always started at zero. The key states found are shown in Table \ref{tab:desc}, which summarizes the key state indices, the quantized encodings, and the decoded actions. It was found the action at the beginning of each episode was almost zero, while the actions associated with other key states were always at the control saturation. More about this phenomenon will be discussed later in this section.

\begin{table}
  \begin{center}
    \caption{Key state description}
    \label{tab:desc}
    \begin{tabular}{|c|c|c|}
      \hline
      Key state index & Quantized encoding & Decoded action\\
      \hline
      $0$ & $[0,~~0,~~0,~~0]^T$ & $[0.03,-0.15,~0.07]^T$ \\
      \hline
      $1$ & $[1,~1,-1,~~1]^T$ & $[2,~~~2,~~~2]^T$ \\
      \hline
      $2$ & $[1,-1,~~1,-1]^T$ & $[-2,~~2,-2]^T$ \\
      \hline
      $3$ & $[-1,-1,~1,-1]^T$ & $[-2,-2,-2]^T$ \\
      \hline
      $4$ & $[-1,~1,-1,~1]^T$ & $[2,~-2,~~2]^T$ \\
      \hline
      $5$ & $[-1,-1,-1,-1]^T$ & $[-2,-2,~~2]^T$ \\
      \hline
    \end{tabular}
  \end{center}
\end{table}

\begin{figure*}
	\centering
	\subfigure[Episode 1]{
		\includegraphics[width=0.3\hsize]{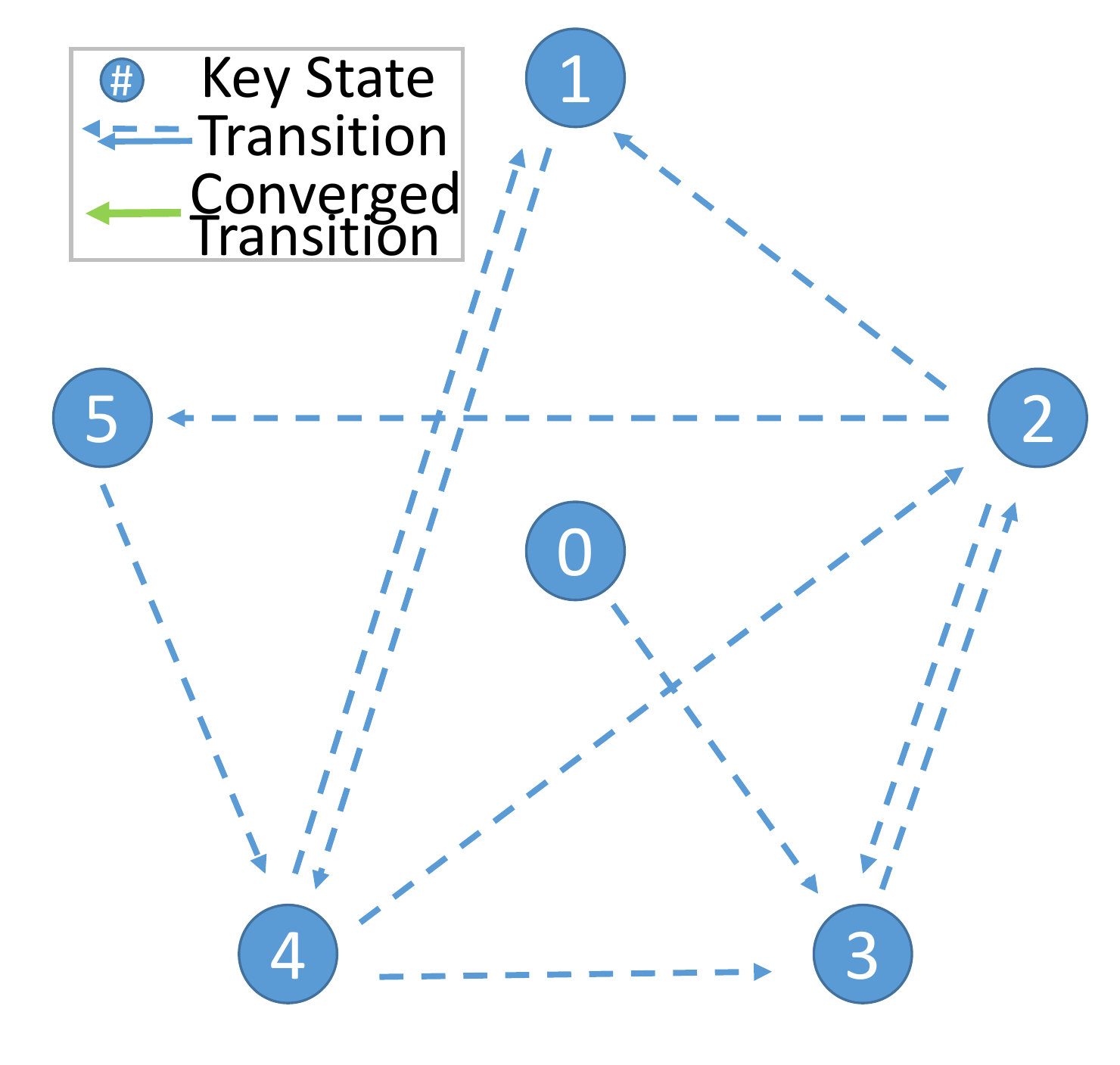}
	}
	\subfigure[Episode 2]{
		\includegraphics[width=0.3\hsize]{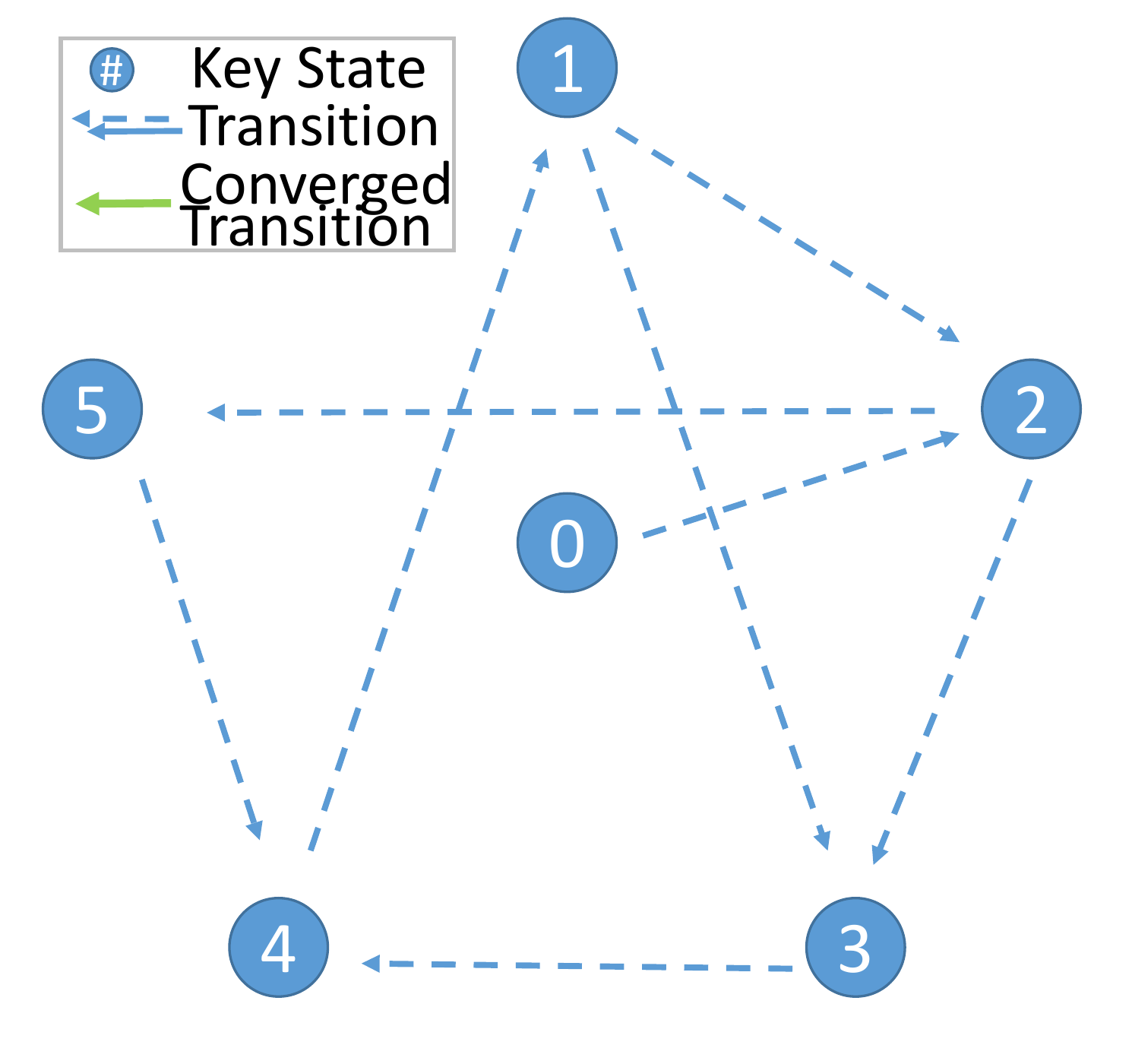}
	}\\
	\subfigure[Episode 3]{
		\includegraphics[width=0.3\hsize]{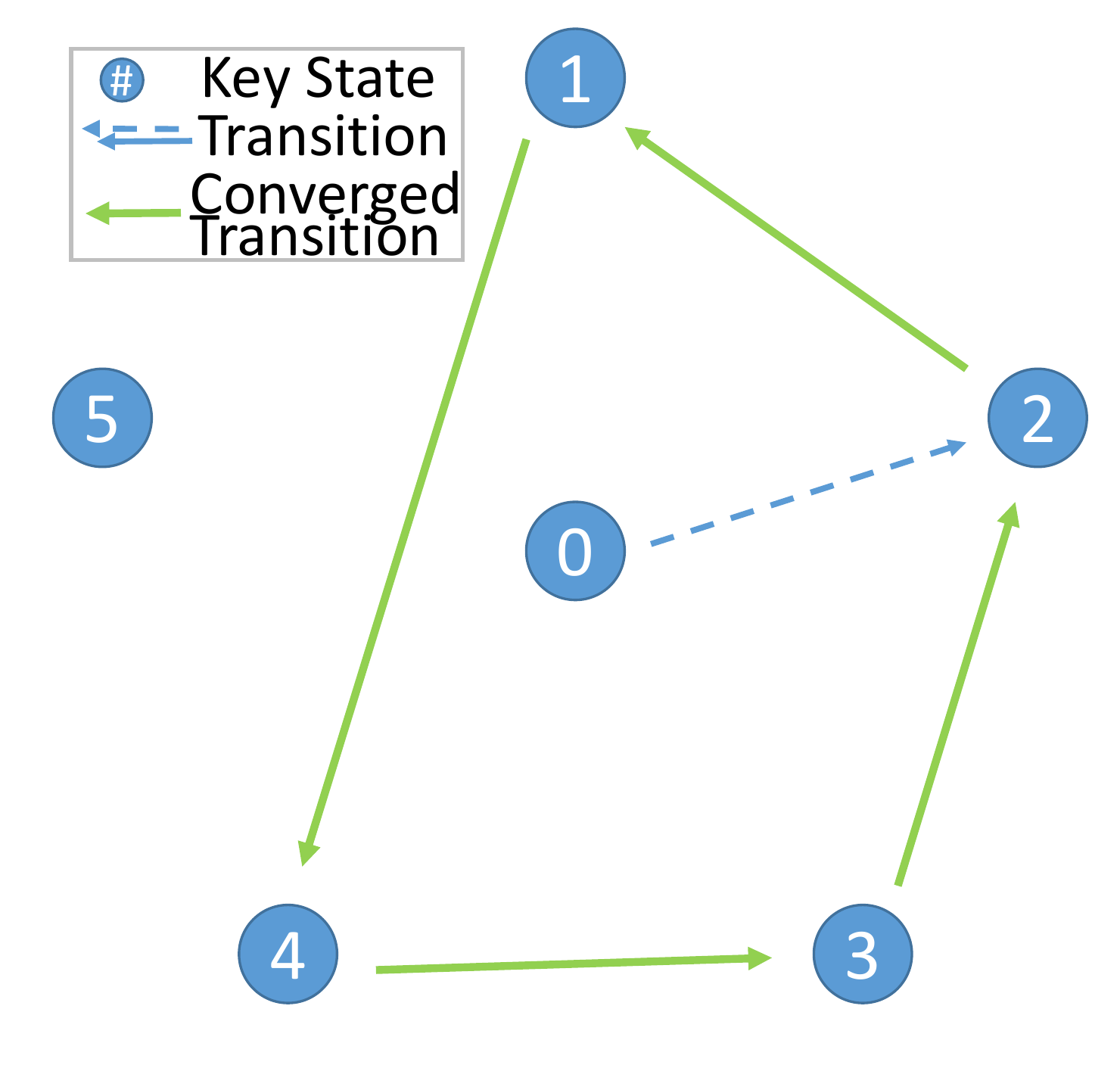}
	}
	\subfigure[Episode 4]{
		\includegraphics[width=0.3\hsize]{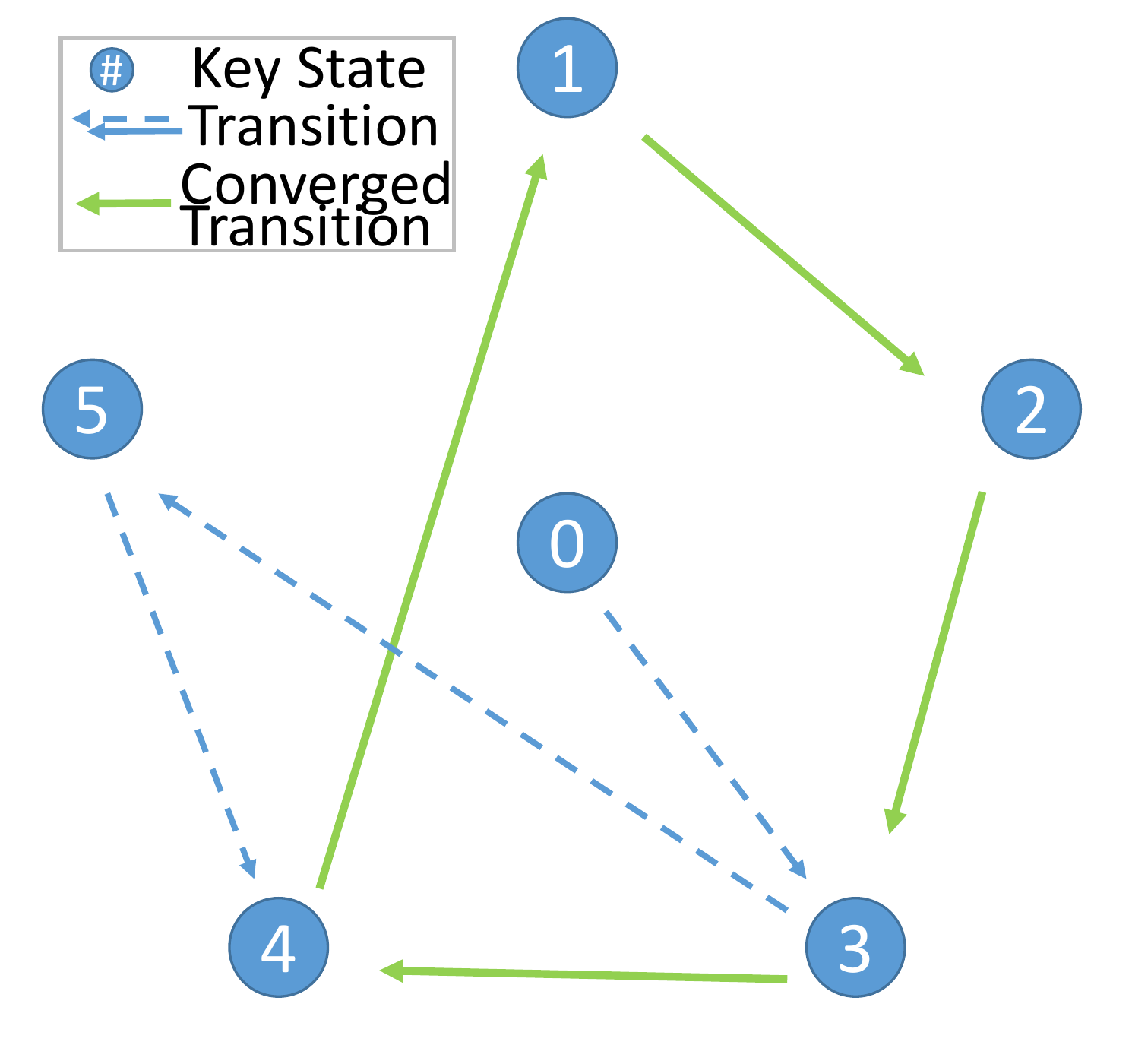}
	}\\
    \subfigure[Episode 5]{
		\includegraphics[width=0.3\hsize]{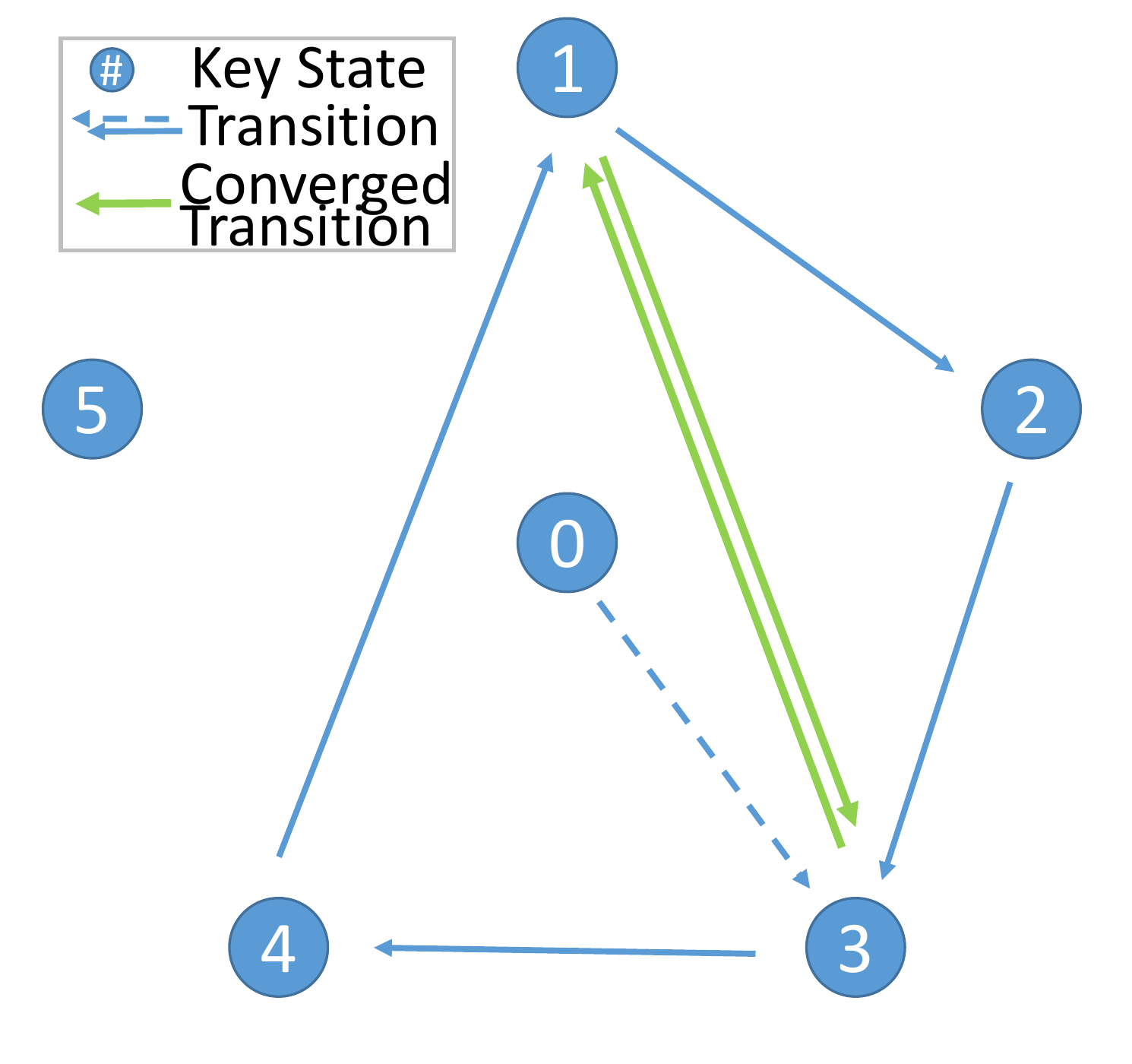}
	}
	\subfigure[Episode 6]{
		\includegraphics[width=0.3\hsize]{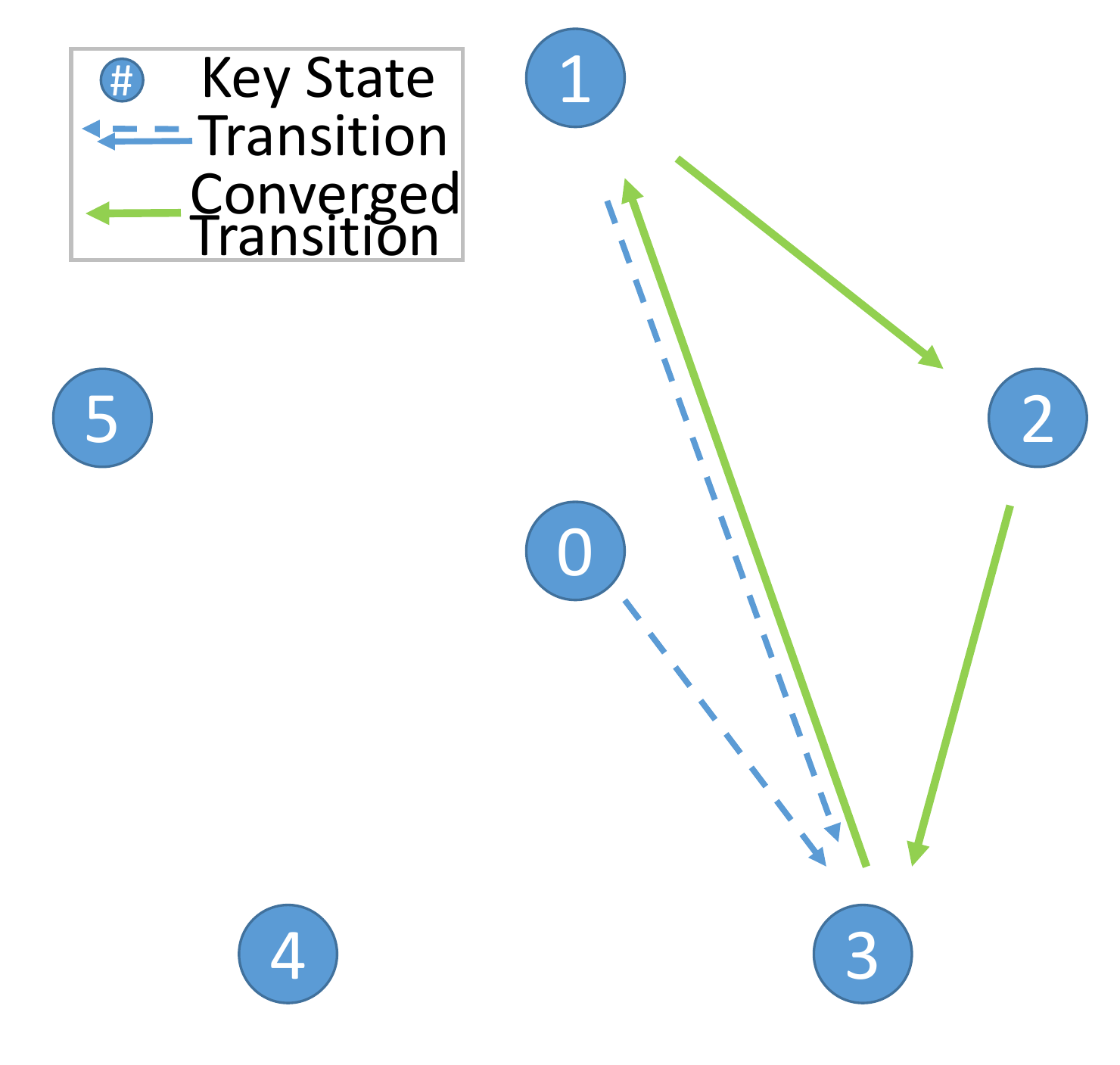}
	}\\
	\subfigure[Episode 7]{
		\includegraphics[width=0.3\hsize]{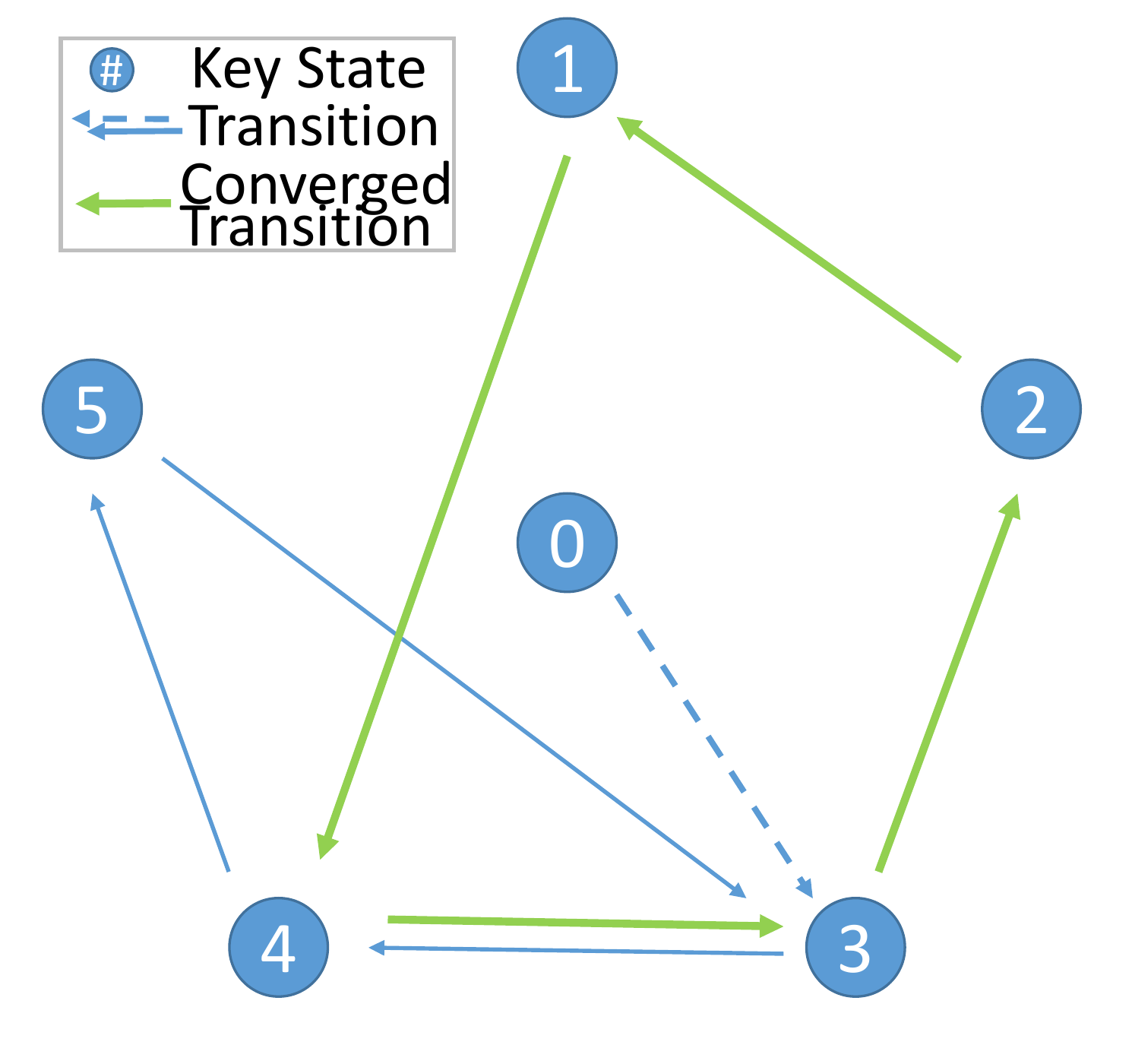}
	}
	\subfigure[Episode 8]{
		\includegraphics[width=0.3\hsize]{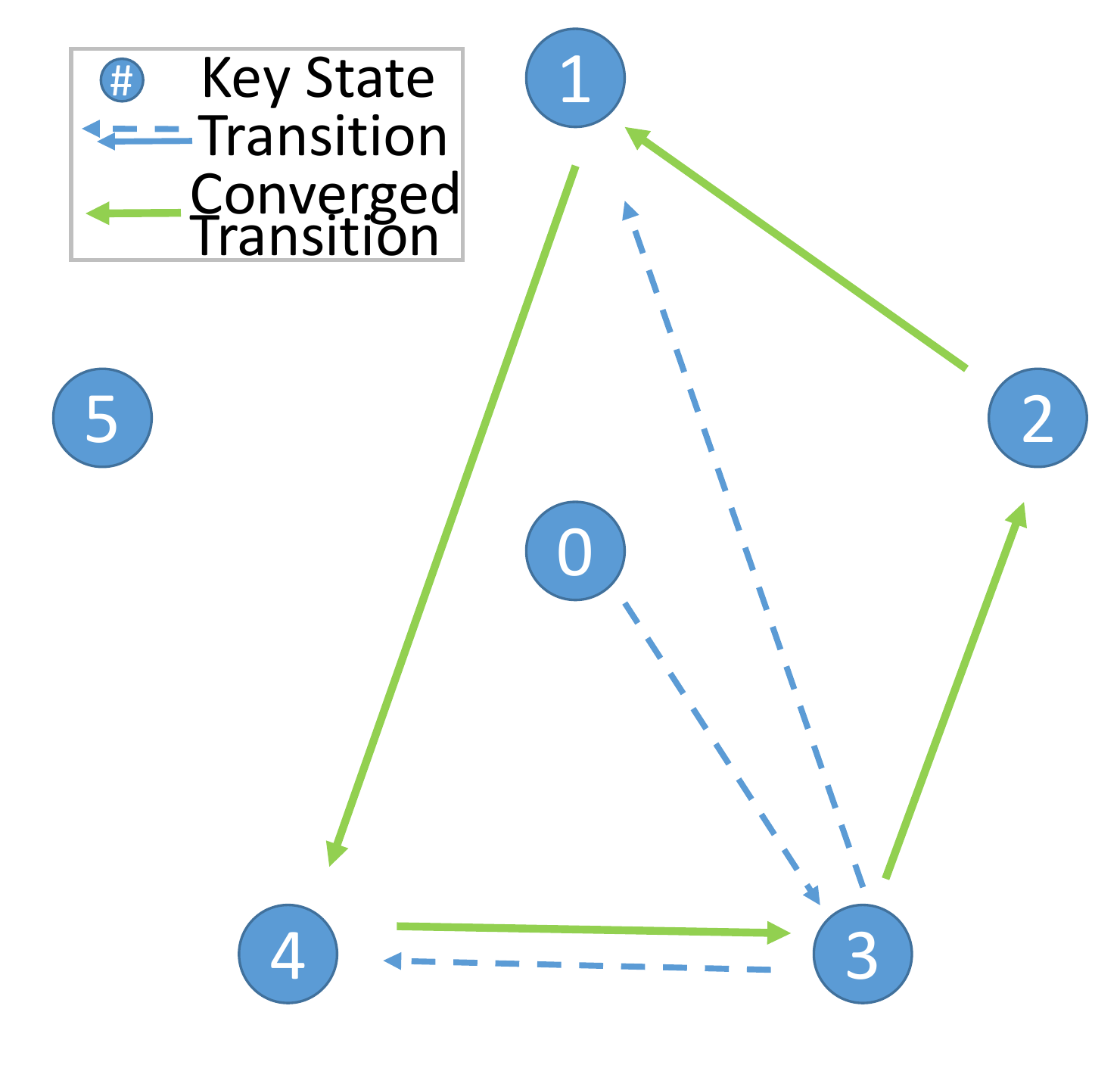}
	}
	\caption{Transitions between key states and cyclic patterns.}
	\label{fig:patterns}
\end{figure*}

The transitions between the key states in $14$ episodes (out of $20$) converged to some cyclic patterns shown in Fig. \ref{fig:patterns}. Figure \ref{fig:patterns} show $8$ examples, where first $2$ examples did not exhibit clear converged patterns. The remaining $6$ examples exhibited three cyclic transition patterns, highlighted by green solid arrows. In all examples, the state started from State $0$ and the system took a number of transitions to enter one of the cyclic patterns. It is because that at the beginning of each simulation (episode), the DOB-net intended to interact with the environments to gain observations for estimating the key aspects of the inherent POMDPs (i.e., disturbances and their transfer functions). The following analysis partially reveal how the hidden states are related to controls and disturbances.

\begin{figure*}
	\centering
	\subfigure[Episode 1]{
		\includegraphics[width=0.4\hsize]{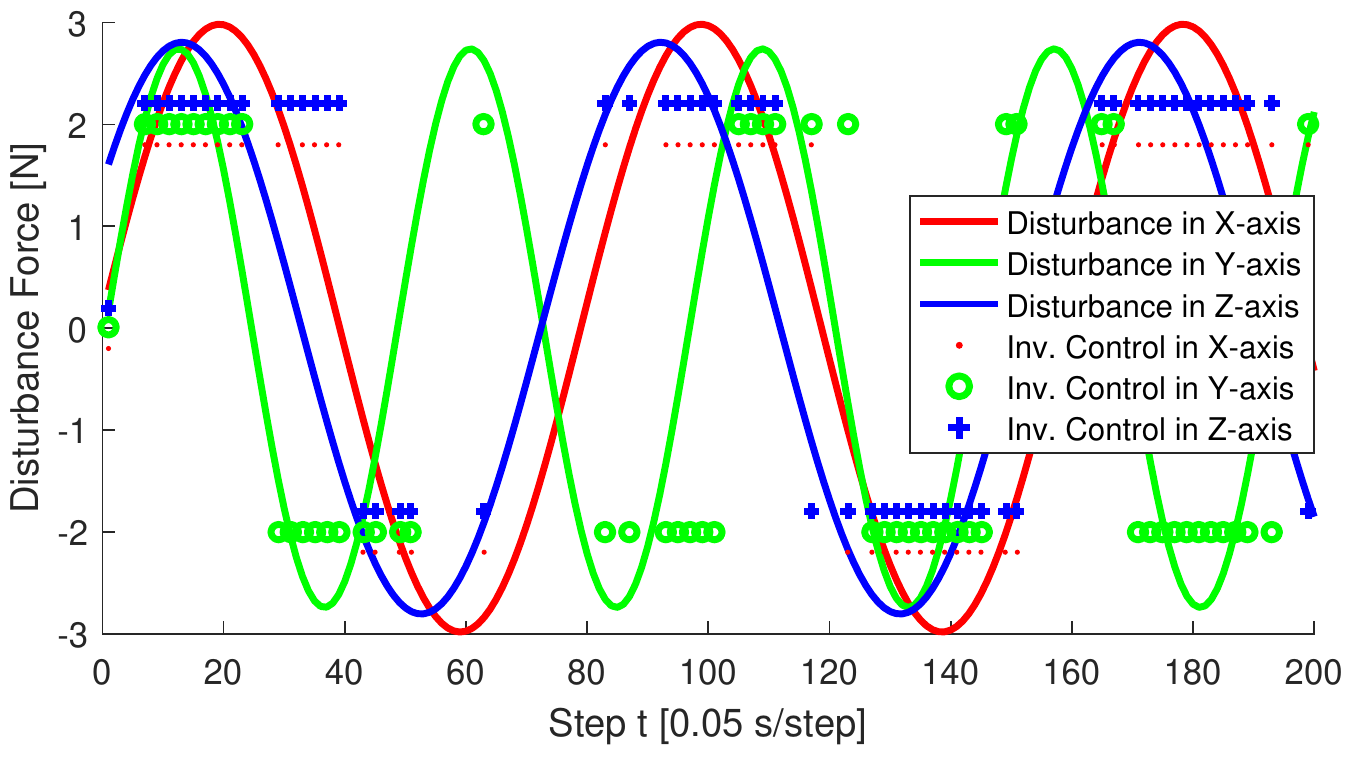}
	}
	\subfigure[Episode 2]{
		\includegraphics[width=0.4\hsize]{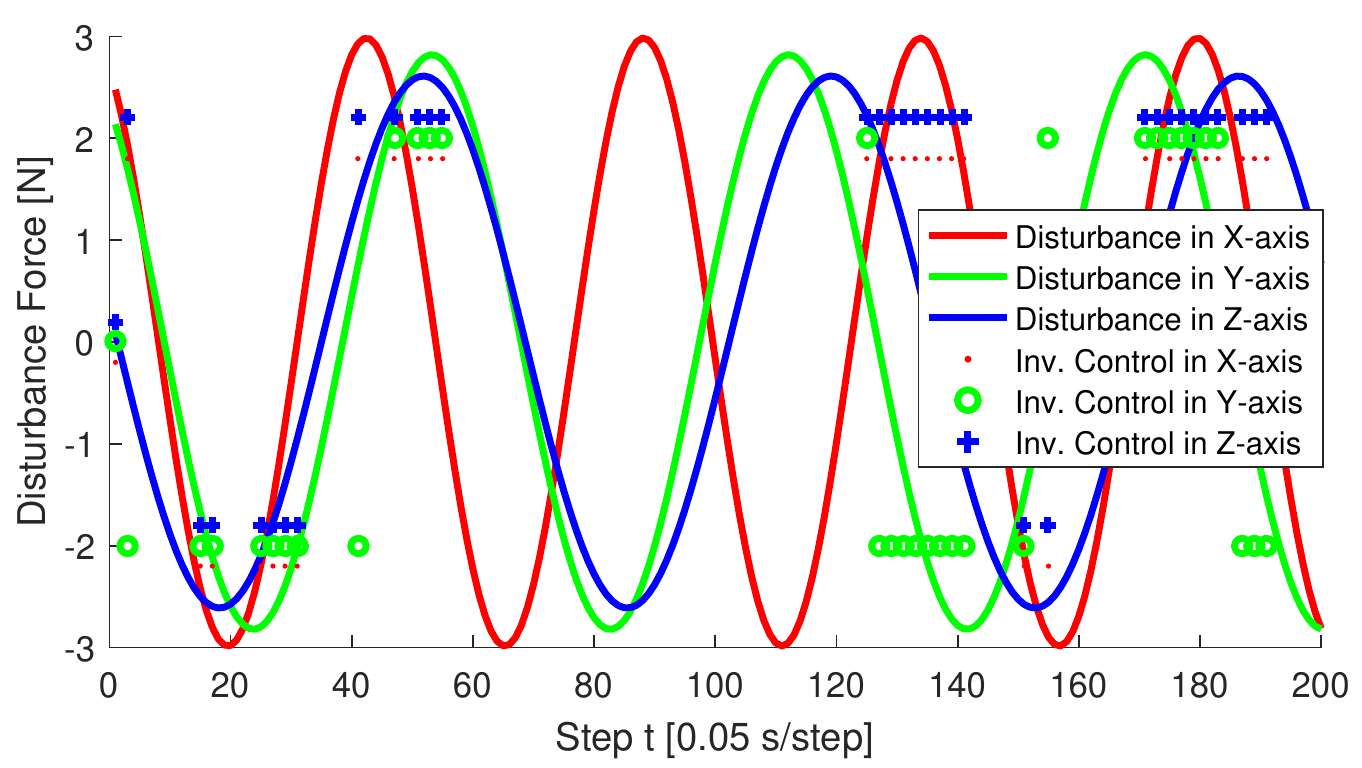}
	}
	\subfigure[Episode 3]{
		\includegraphics[width=0.4\hsize]{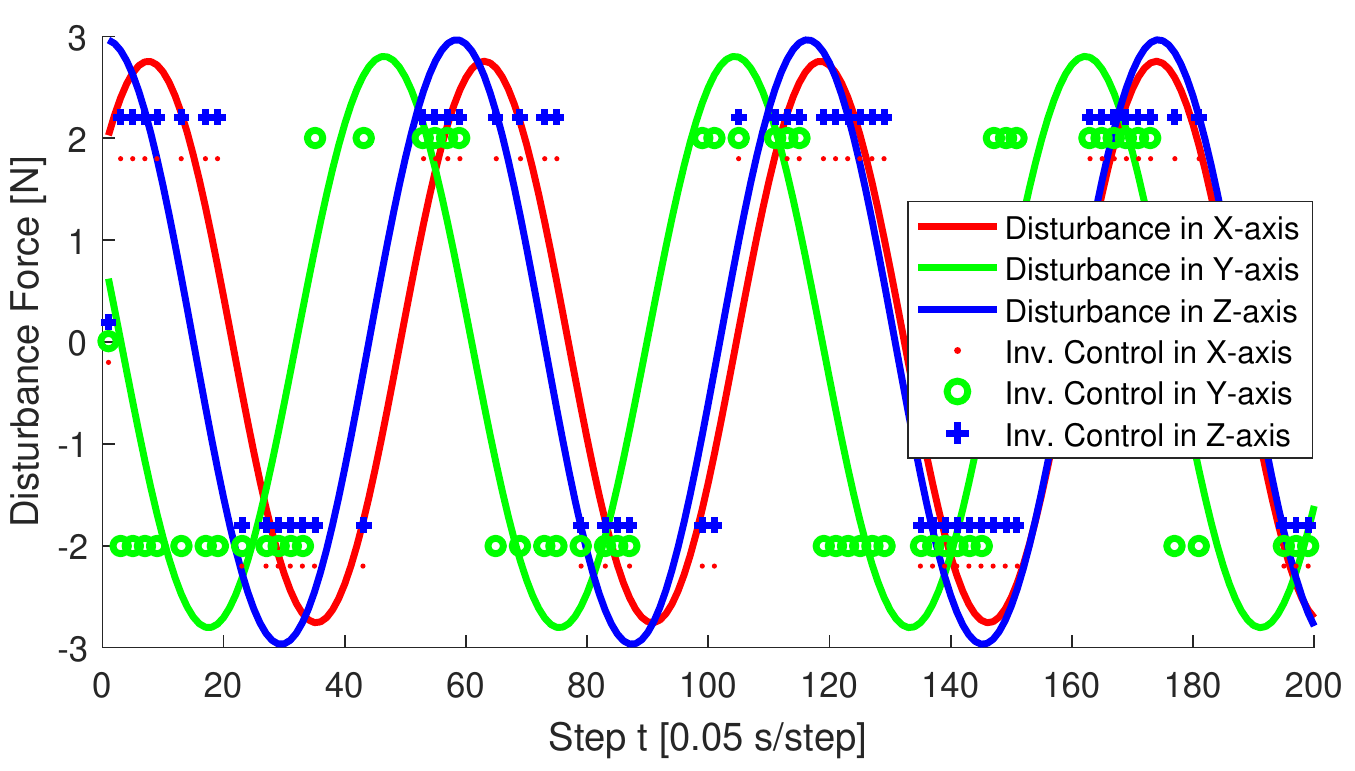}
	}
	\subfigure[Episode 4]{
		\includegraphics[width=0.4\hsize]{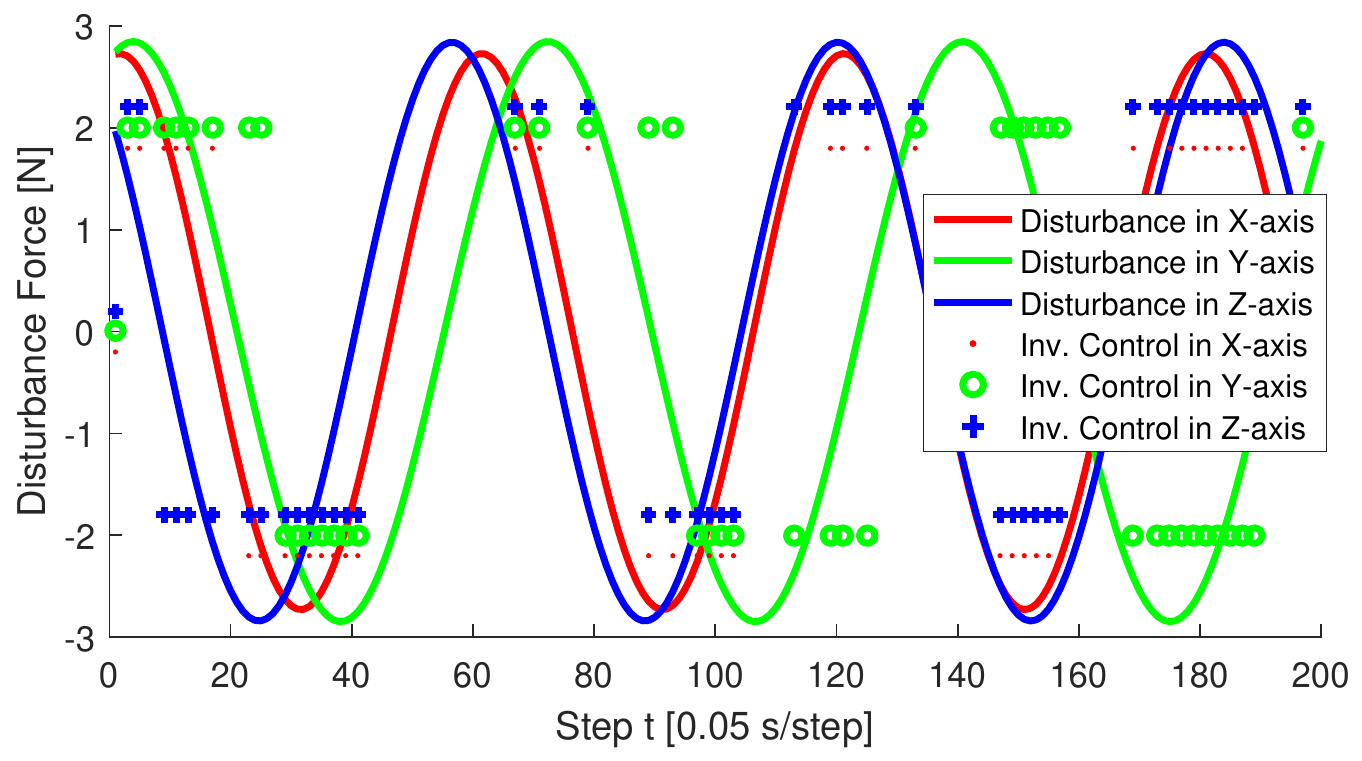}
	}
	\hspace{0.2cm}
    \subfigure[Episode 5]{
		\includegraphics[width=0.4\hsize]{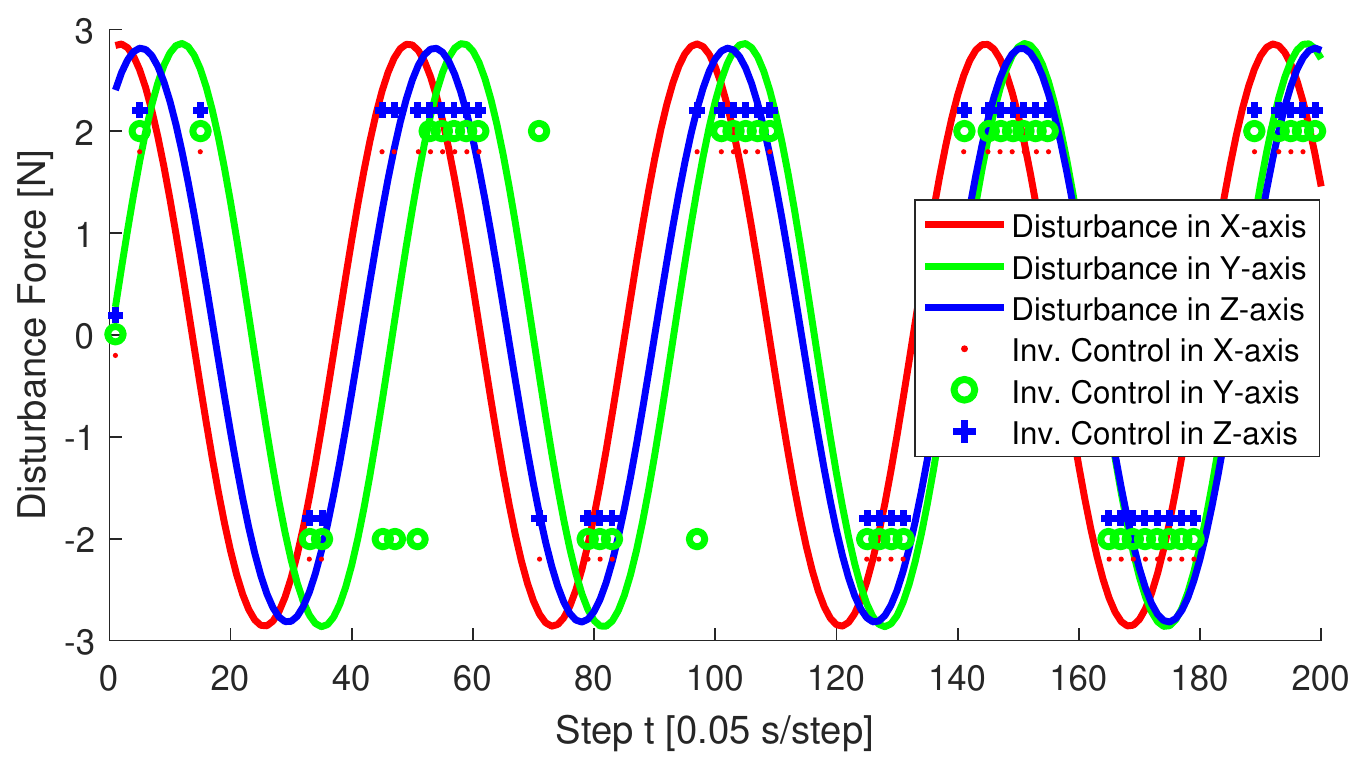}
	}
	\subfigure[Episode 6]{
		\includegraphics[width=0.4\hsize]{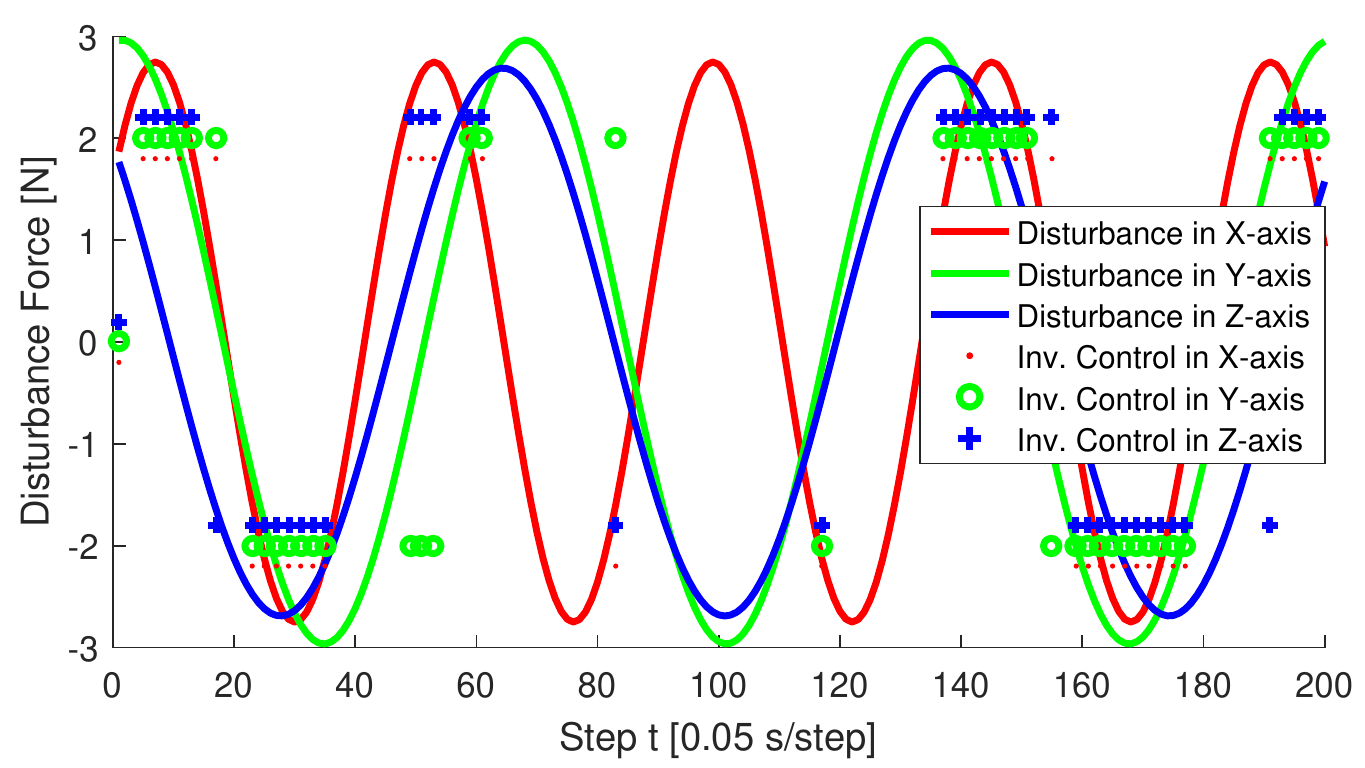}
	}
	\subfigure[Episode 7]{
		\includegraphics[width=0.4\hsize]{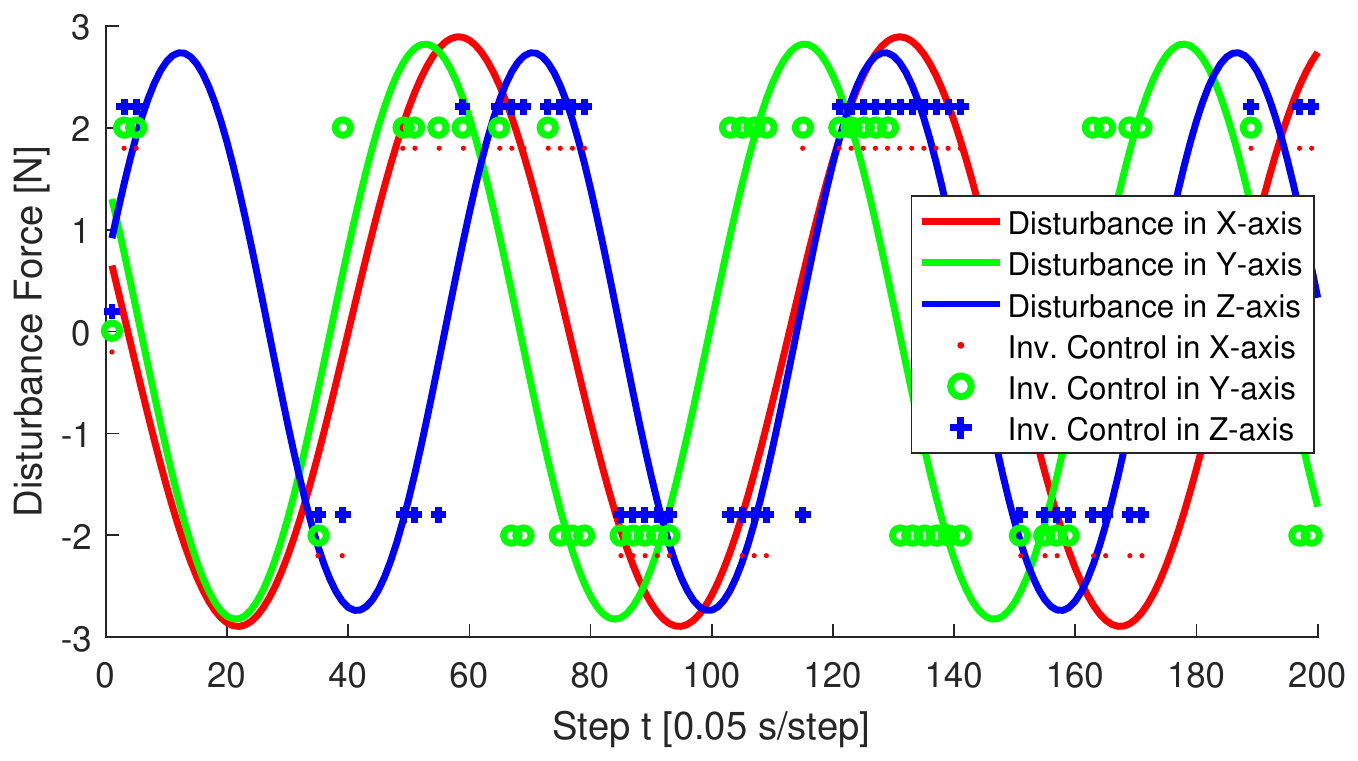}
	}
	\subfigure[Episode 8]{
		\includegraphics[width=0.4\hsize]{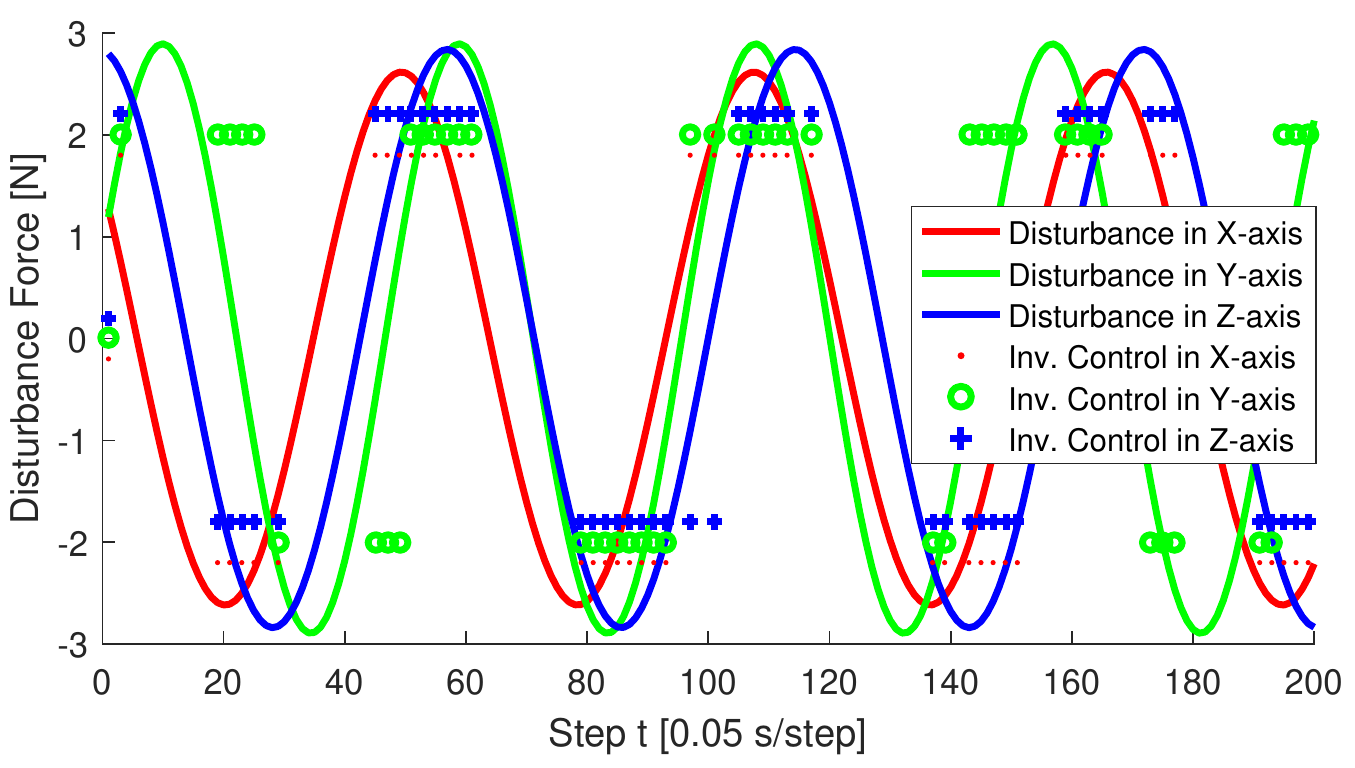}
	}
	\caption{Disturbances and additive inverse controls at key states.}
	\label{fig:disturbances}
\end{figure*}

Considering the associated action with each state in the KMMN, it was found that the learnt DOB-net behaved similarly to a hybrid controller where switchings occur. These switchings exhibited cyclic patterns due to the fact the disturbance in each direction was periodic. Each switching pattern indicated a disturbance pattern. As shown in Fig. \ref{fig:disturbances}, the disturbances in three directions are illustrated in red, green, and blue, respectively. The additive inversion of the controls associated with the states are also illustrated. Note that the values of the controls in $x$ and $z$ directions were added by $-0.2$ and $0.2$, respectively, for clear illustration. It was found that the states in the KMMN were only activated when the disturbances were close to the control saturation, as shown in Fig. \ref{fig:disturbances}. By inspecting the controls and unknown disturbances, it was shown that the obtained actions were synchronized with the disturbance forces.

Some episodes exhibited similar converged transition patterns, as shown in Fig. \ref{fig:patterns} (c), (d), (f), and (h). However, the way the system entered into the cyclic patterns varied in different environments. The system shown in Fig. \ref{fig:patterns}(c) entered the cyclic pattern (referred to as cycle) through State $2$ directly, while the system shown in Fig. \ref{fig:patterns}(h) entered the cycle through State $4$ after visiting State $3$. In Fig. \ref{fig:patterns}(d), the system visited States $3$ and $5$, and then entered into the cycle at State $4$. As illustrated in Fig. \ref{fig:disturbances}, in cases of (c), (d), (f), and (h), the disturbance forces in $x$ and $y$ directions had similar frequencies and phases, while the disturbance force in $z$ direction had different a frequency and phase. These examples show that the DOB-net was able to estimate disturbances and their inherent governing behavior.

In Fig. \ref{fig:patterns}(e) and (f), the systems exhibited another two cyclic patterns. Interestingly, the episodes shown in Fig. \ref{fig:patterns}(g) and (e) exhibited two cyclic patterns, respectively. For example, the system in Fig. \ref{fig:patterns}(g) first entered the cycle (shown in blue solid lines) through State $3$ and then entered the second cycle at State $4$ and stayed in the second cycle.

The first two examples in Fig. \ref{fig:patterns} did not exhibit clear cyclic patterns. It is possible that the key states found in those $20$ episodes did not capture the states that were crucial to this two examples. More research about the definition of sufficient attention should be explored in future research.

The system in Fig. \ref{fig:patterns}(e) entered in a binary switching pattern. With careful examination of the disturbances in Fig. \ref{fig:disturbances}(e), we found the components in the randomly-generated disturbances had similar periods and phases. Therefore, the two states in the KMMN were sufficient to capture the periodic shifts.
Overall, the key states found have strong correlation between disturbance patterns and the time instants when the disturbance forces were close to control saturation. The phases between disturbances change as a function of time, as shown in Fig. \ref{fig:disturbances}, which strongly ties the change of the hidden states and the action associated. Therefore, the observer designed in the DOB-net and learned together with the control subnetwork was able to estimate such shift in the phases and magnitudes of the disturbances.

\section{DISCUSSION}
\label{sec:discuss}
\begin{figure}
	\centering \includegraphics[width=0.7\hsize]{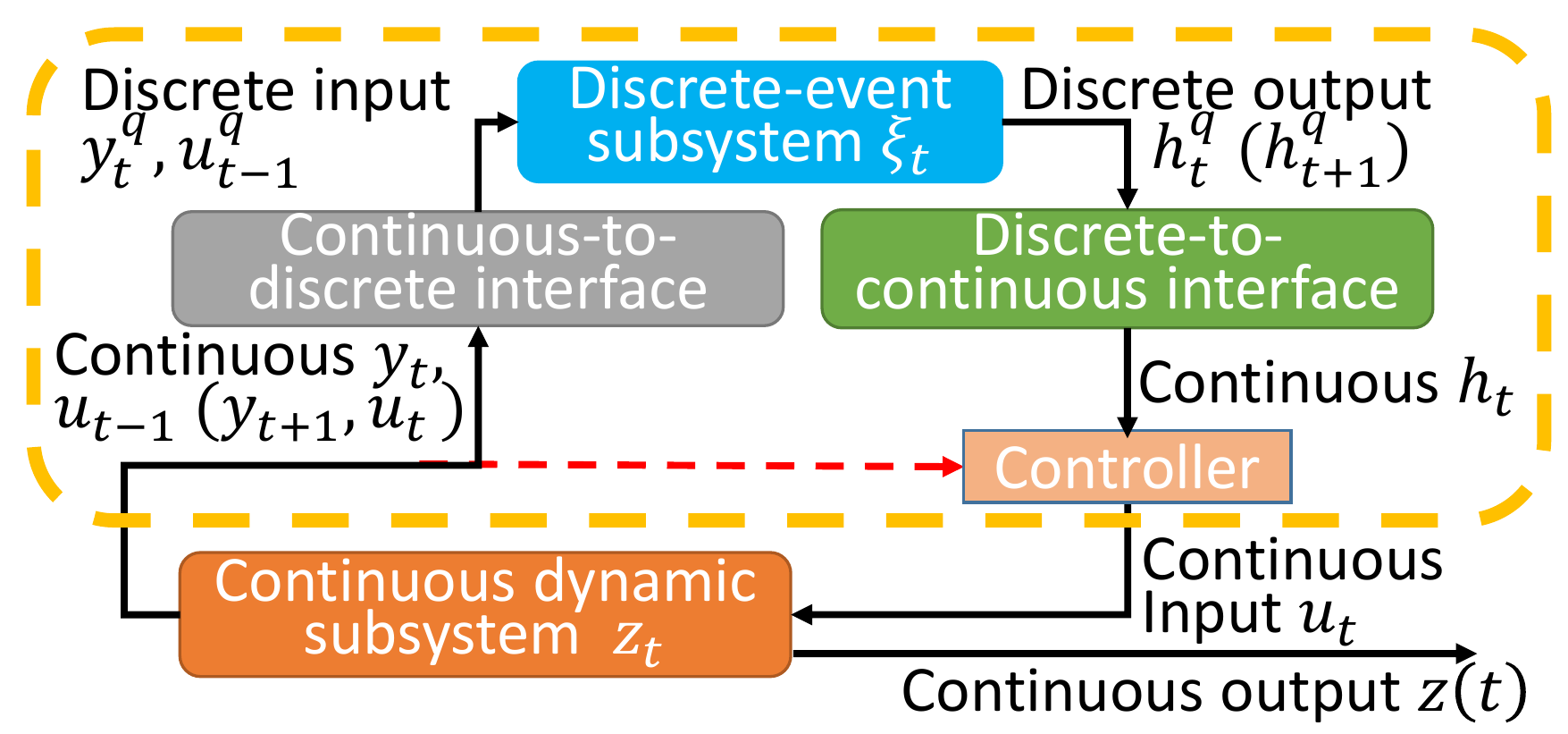}
	\caption{Hybrid structure of the controlled often-saturated system. Except the red dashed arrow to the controller, components within dash-rounded rectangle is equivalent to the DOB-net. }
	\label{fig:hybrid_system}
\end{figure}

As pointed in \cite{zuo2010fault, benzaouia2010stabilisation, yuan2015switching, dong2010model}, the controlled platform whose control often reaches control saturation can be described by a switching-control-regulated system. This kind of systems can be characterized by
\begin{align}
\xi(t+1) &= \sigma \big( \xi(t), y(t) \big)\nonumber\\
z(t+1) &= \eta_{\xi(t)} \big(z(t),u(t) \big)\nonumber,
\end{align}
where $\xi(k+1)$ is discrete state, $\sigma(\cdot)$ governs the switching between the discrete states (refer to as ``modes" in hybrid control), $\eta(\cdot)$ defines the transition function of the continuous state $x(k)$. Then the controlled platform can be depicted as the structure in Fig. \ref{fig:hybrid_system}.

The relation between the quantized DOB-net and the hybrid-system control can be found by comparing Figures \ref{fig:hybrid_system} and \ref{fig:dob_quantized}. The components in the blue dashed rectangle in Fig. \ref{fig:dob_quantized} correspond to the discrete-event subsystem, which is represented as the blue rounded rectangle in Fig. \ref{fig:hybrid_system}. The components in the green dash-dotted rectangle in Fig. \ref{fig:dob_quantized} correspond to the mapping between the discrete hidden states and the continuous controls, i.e., the discrete-to-continuous interface and the controller shown in Fig. \ref{fig:hybrid_system}.

Note that in classical hybrid modelling and control, the red dashed arrow to the controller is necessary, which is not kept in the quantized DOB-net. Therefore, the quantized DOB-net only captures the discrete-event subsystem, which partially describes the interplay between the control strategy and the environments. The DOB-net is able to estimate the discrete-event subsystem online and generate its sufficient representation for effective control.

Cyclic switchings were found in the learnt DOB-net, showing the control policy is able to capture $\sigma(\cdot)$ for position regulation problem in different environments (different POMDPs). In Fig. \ref{fig:disturbances}, the control between switching were not depicted for clear illustration, which may reflect $\eta_{\xi}(\cdot)$. The continuous control based on feedback from continuous observation is missing in this study and should be included for future research.

\section{CONCLUSION \& FUTURE WORK}
\label{sec:conclusion}
This paper proposes an attention-based abstraction approach for finding a key Moore machine network, which reveals the switching mechanism that has been captured in the DOB-net and is key to excessive disturbance rejection. This method is effective in abstracting control logics in solving different POMDPs. Interestingly, the switching mechanism has been manually designed for controller developments in existing literature. This finding may offer a bridge between DOB-nets and the hybrid systems for better analysis.

In the future, more effort will be devoted to a new definition of sufficient attention to better capture the control mechanisms common in solving multiple POMDPs. Also, the continuous controls should be characterized to show how the system is guided between switchings, for the purpose of fully understanding the control network in the language of hybrid control. Another interesting future work is to investigate the possibility of using the switching mechanism obtained through inductive learning as some distilled knowledge for transfer learning.


\bibliographystyle{cas-model2-names}

\bibliography{./bibliographies/mybibfile,./bibliographies/Other,./bibliographies/WenjieLuPaper}

\begin{thebibliography}{61}
\expandafter\ifx\csname natexlab\endcsname\relax\def\natexlab#1{#1}\fi
\providecommand{\url}[1]{\texttt{#1}}
\providecommand{\href}[2]{#2}
\providecommand{\path}[1]{#1}
\providecommand{\DOIprefix}{doi:}
\providecommand{\ArXivprefix}{arXiv:}
\providecommand{\URLprefix}{URL: }
\providecommand{\Pubmedprefix}{pmid:}
\providecommand{\doi}[1]{\href{http://dx.doi.org/#1}{\path{#1}}}
\providecommand{\Pubmed}[1]{\href{pmid:#1}{\path{#1}}}
\providecommand{\bibinfo}[2]{#2}
\ifx\xfnm\relax \def\xfnm[#1]{\unskip,\space#1}\fi
\bibitem[{{\AA}str{\"o}m and Wittenmark(2013)}]{aastrom2013adaptive}
\bibinfo{author}{{\AA}str{\"o}m, K.J.}, \bibinfo{author}{Wittenmark, B.},
  \bibinfo{year}{2013}.
\newblock \bibinfo{title}{Adaptive control}.
\newblock \bibinfo{publisher}{Courier Corporation}.
\bibitem[{Barnsley(2014)}]{barnsley2014fractals}
\bibinfo{author}{Barnsley, M.F.}, \bibinfo{year}{2014}.
\newblock \bibinfo{title}{Fractals everywhere}.
\newblock \bibinfo{publisher}{Academic press}.
\bibitem[{Barreto and et~al.(2017)}]{barreto2017successor}
\bibinfo{author}{Barreto, A.}, \bibinfo{author}{et~al.}, \bibinfo{year}{2017}.
\newblock \bibinfo{title}{Successor features for transfer in reinforcement
  learning}, in: \bibinfo{booktitle}{Advances in neural information processing
  systems}, pp. \bibinfo{pages}{4055--4065}.
\bibitem[{Bengio et~al.(2013)Bengio, L{\'e}onard and
  Courville}]{bengio2013estimating}
\bibinfo{author}{Bengio, Y.}, \bibinfo{author}{L{\'e}onard, N.},
  \bibinfo{author}{Courville, A.}, \bibinfo{year}{2013}.
\newblock \bibinfo{title}{Estimating or propagating gradients through
  stochastic neurons for conditional computation}.
\newblock \bibinfo{journal}{arXiv preprint arXiv:1308.3432} .
\bibitem[{Benzaouia et~al.(2010)Benzaouia, Akhrif and
  Saydy}]{benzaouia2010stabilisation}
\bibinfo{author}{Benzaouia, A.}, \bibinfo{author}{Akhrif, O.},
  \bibinfo{author}{Saydy, L.}, \bibinfo{year}{2010}.
\newblock \bibinfo{title}{Stabilisation and control synthesis of switching
  systems subject to actuator saturation}.
\newblock \bibinfo{journal}{International Journal of Systems Science}
  \bibinfo{volume}{41}, \bibinfo{pages}{397--409}.
\bibitem[{Brahmbhatt and Hays(2017)}]{brahmbhatt2017deepnav}
\bibinfo{author}{Brahmbhatt, S.}, \bibinfo{author}{Hays, J.},
  \bibinfo{year}{2017}.
\newblock \bibinfo{title}{Deepnav: Learning to navigate large cities}, in:
  \bibinfo{booktitle}{Proc. IEEE Conf. Comput. Vis. Pattern Recognit.}, pp.
  \bibinfo{pages}{3087--3096}.
\bibitem[{Camacho and Alba(2013)}]{camacho2013model}
\bibinfo{author}{Camacho, E.F.}, \bibinfo{author}{Alba, C.B.},
  \bibinfo{year}{2013}.
\newblock \bibinfo{title}{Model predictive control}.
\newblock \bibinfo{publisher}{Springer Science \& Business Media}.
\bibitem[{Chen et~al.(2000)Chen, Ballance, Gawthrop and
  O'Reilly}]{chen2000nonlinear}
\bibinfo{author}{Chen, W.H.}, \bibinfo{author}{Ballance, D.J.},
  \bibinfo{author}{Gawthrop, P.J.}, \bibinfo{author}{O'Reilly, J.},
  \bibinfo{year}{2000}.
\newblock \bibinfo{title}{A nonlinear disturbance observer for robotic
  manipulators}.
\newblock \bibinfo{journal}{IEEE Transactions on industrial Electronics}
  \bibinfo{volume}{47}, \bibinfo{pages}{932--938}.
\bibitem[{Cheng and Krishnakumar(1993)}]{cheng1993automatic}
\bibinfo{author}{Cheng, K.T.}, \bibinfo{author}{Krishnakumar, A.S.},
  \bibinfo{year}{1993}.
\newblock \bibinfo{title}{Automatic functional test generation using the
  extended finite state machine model}, in: \bibinfo{booktitle}{30th ACM/IEEE
  Design Automation Conference}, \bibinfo{organization}{IEEE}. pp.
  \bibinfo{pages}{86--91}.
\bibitem[{Cleeremans et~al.(1989)Cleeremans, Servan-Schreiber and
  McClelland}]{cleeremans1989finite}
\bibinfo{author}{Cleeremans, A.}, \bibinfo{author}{Servan-Schreiber, D.},
  \bibinfo{author}{McClelland, J.L.}, \bibinfo{year}{1989}.
\newblock \bibinfo{title}{Finite state automata and simple recurrent networks}.
\newblock \bibinfo{journal}{Neural computation} \bibinfo{volume}{1},
  \bibinfo{pages}{372--381}.
\bibitem[{Crutchfield and Young(1988)}]{crutchfield1988computation}
\bibinfo{author}{Crutchfield, J.P.}, \bibinfo{author}{Young, K.},
  \bibinfo{year}{1988}.
\newblock \bibinfo{title}{Computation at the onset of chaos}, in:
  \bibinfo{booktitle}{The Santa Fe Institute, Westview},
  \bibinfo{organization}{Citeseer}.
\bibitem[{Dong et~al.(2010)Dong, Hou, Zhang and Wang}]{dong2010model}
\bibinfo{author}{Dong, C.}, \bibinfo{author}{Hou, Y.}, \bibinfo{author}{Zhang,
  Y.}, \bibinfo{author}{Wang, Q.}, \bibinfo{year}{2010}.
\newblock \bibinfo{title}{Model reference adaptive switching control of a
  linearized hypersonic flight vehicle model with actuator saturation}.
\newblock \bibinfo{journal}{Proceedings of the Institution of Mechanical
  Engineers, Part I: Journal of Systems and Control Engineering}
  \bibinfo{volume}{224}, \bibinfo{pages}{289--303}.
\bibitem[{Doyle et~al.(1989)Doyle, Glover, Khargonekar and
  Francis}]{doyle1989state}
\bibinfo{author}{Doyle, J.C.}, \bibinfo{author}{Glover, K.},
  \bibinfo{author}{Khargonekar, P.P.}, \bibinfo{author}{Francis, B.A.},
  \bibinfo{year}{1989}.
\newblock \bibinfo{title}{State-space solutions to standard h/sub 2/and h/sub
  infinity/control problems}.
\newblock \bibinfo{journal}{IEEE Transactions on Automatic control}
  \bibinfo{volume}{34}, \bibinfo{pages}{831--847}.
\bibitem[{Edwards and Spurgeon(1998)}]{edwards1998sliding}
\bibinfo{author}{Edwards, C.}, \bibinfo{author}{Spurgeon, S.},
  \bibinfo{year}{1998}.
\newblock \bibinfo{title}{Sliding mode control: theory and applications}.
\newblock \bibinfo{publisher}{Crc Press}.
\bibitem[{Frasconi et~al.(1996)Frasconi, Gori, Maggini and
  Soda}]{frasconi1996representation}
\bibinfo{author}{Frasconi, P.}, \bibinfo{author}{Gori, M.},
  \bibinfo{author}{Maggini, M.}, \bibinfo{author}{Soda, G.},
  \bibinfo{year}{1996}.
\newblock \bibinfo{title}{Representation of finite state automata in recurrent
  radial basis function networks}.
\newblock \bibinfo{journal}{Machine Learning} \bibinfo{volume}{23},
  \bibinfo{pages}{5--32}.
\bibitem[{Gao and Cai(2016)}]{gao2016nonlinear}
\bibinfo{author}{Gao, H.}, \bibinfo{author}{Cai, Y.}, \bibinfo{year}{2016}.
\newblock \bibinfo{title}{Nonlinear disturbance observer-based model predictive
  control for a generic hypersonic vehicle}.
\newblock \bibinfo{journal}{Proceedings of the Institution of Mechanical
  Engineers, Part I: Journal of Systems and Control Engineering}
  \bibinfo{volume}{230}, \bibinfo{pages}{3--12}.
\bibitem[{Gao(2014)}]{gao2014centrality}
\bibinfo{author}{Gao, Z.}, \bibinfo{year}{2014}.
\newblock \bibinfo{title}{On the centrality of disturbance rejection in
  automatic control}.
\newblock \bibinfo{journal}{ISA transactions} \bibinfo{volume}{53},
  \bibinfo{pages}{850--857}.
\bibitem[{Gao et~al.(2001)Gao, Huang and Han}]{gao2001alternative}
\bibinfo{author}{Gao, Z.}, \bibinfo{author}{Huang, Y.}, \bibinfo{author}{Han,
  J.}, \bibinfo{year}{2001}.
\newblock \bibinfo{title}{An alternative paradigm for control system design},
  in: \bibinfo{booktitle}{Decision and Control, 2001. Proceedings of the 40th
  IEEE Conference on}, \bibinfo{organization}{IEEE}. pp.
  \bibinfo{pages}{4578--4585}.
\bibitem[{Ghafarirad et~al.(2014)Ghafarirad, Rezaei, Zareinejad and
  Sarhan}]{ghafarirad2014disturbance}
\bibinfo{author}{Ghafarirad, H.}, \bibinfo{author}{Rezaei, S.M.},
  \bibinfo{author}{Zareinejad, M.}, \bibinfo{author}{Sarhan, A.A.},
  \bibinfo{year}{2014}.
\newblock \bibinfo{title}{Disturbance rejection-based robust control for
  micropositioning of piezoelectric actuators}.
\newblock \bibinfo{journal}{Comptes Rendus M{\'e}canique}
  \bibinfo{volume}{342}, \bibinfo{pages}{32--45}.
\bibitem[{Gu et~al.(2016a)Gu, Lillicrap, Ghahramani, Turner and
  Levine}]{gu2016q}
\bibinfo{author}{Gu, S.}, \bibinfo{author}{Lillicrap, T.},
  \bibinfo{author}{Ghahramani, Z.}, \bibinfo{author}{Turner, R.E.},
  \bibinfo{author}{Levine, S.}, \bibinfo{year}{2016}a.
\newblock \bibinfo{title}{Q-prop: Sample-efficient policy gradient with an
  off-policy critic}.
\newblock \bibinfo{journal}{arXiv preprint arXiv:1611.02247} .
\bibitem[{Gu et~al.(2016b)Gu, Lillicrap, Sutskever and
  Levine}]{gu2016continuous}
\bibinfo{author}{Gu, S.}, \bibinfo{author}{Lillicrap, T.},
  \bibinfo{author}{Sutskever, I.}, \bibinfo{author}{Levine, S.},
  \bibinfo{year}{2016}b.
\newblock \bibinfo{title}{Continuous deep q-learning with model-based
  acceleration}, in: \bibinfo{booktitle}{International Conference on Machine
  Learning}, pp. \bibinfo{pages}{2829--2838}.
\bibitem[{Gunning(2017)}]{gunning2017explainable}
\bibinfo{author}{Gunning, D.}, \bibinfo{year}{2017}.
\newblock \bibinfo{title}{Explainable artificial intelligence (xai)}.
\newblock \bibinfo{journal}{Defense Advanced Research Projects Agency (DARPA),
  nd Web} \bibinfo{volume}{2}.
\bibitem[{Han(1995)}]{han1995extended}
\bibinfo{author}{Han, J.}, \bibinfo{year}{1995}.
\newblock \bibinfo{title}{The" extended state observer" of a class of uncertain
  systems [j]}.
\newblock \bibinfo{journal}{Control and Decision} \bibinfo{volume}{1}.
\bibitem[{Hinton and Salakhutdinov(2006)}]{hinton2006reducing}
\bibinfo{author}{Hinton, G.E.}, \bibinfo{author}{Salakhutdinov, R.R.},
  \bibinfo{year}{2006}.
\newblock \bibinfo{title}{Reducing the dimensionality of data with neural
  networks}.
\newblock \bibinfo{journal}{science} \bibinfo{volume}{313},
  \bibinfo{pages}{504--507}.
\bibitem[{Hubara et~al.(2016)Hubara, Courbariaux, Soudry, El-Yaniv and
  Bengio}]{hubara2016binarized}
\bibinfo{author}{Hubara, I.}, \bibinfo{author}{Courbariaux, M.},
  \bibinfo{author}{Soudry, D.}, \bibinfo{author}{El-Yaniv, R.},
  \bibinfo{author}{Bengio, Y.}, \bibinfo{year}{2016}.
\newblock \bibinfo{title}{Binarized neural networks}, in:
  \bibinfo{booktitle}{Advances in neural information processing systems}, pp.
  \bibinfo{pages}{4107--4115}.
\bibitem[{Johnson(1968)}]{johnson1968optimal}
\bibinfo{author}{Johnson, C.}, \bibinfo{year}{1968}.
\newblock \bibinfo{title}{Optimal control of the linear regulator with constant
  disturbances}.
\newblock \bibinfo{journal}{IEEE Transactions on Automatic Control}
  \bibinfo{volume}{13}, \bibinfo{pages}{416--421}.
\bibitem[{Johnson(1971)}]{johnson1971accomodation}
\bibinfo{author}{Johnson, C.}, \bibinfo{year}{1971}.
\newblock \bibinfo{title}{Accomodation of external disturbances in linear
  regulator and servomechanism problems}.
\newblock \bibinfo{journal}{IEEE Transactions on automatic control}
  \bibinfo{volume}{16}, \bibinfo{pages}{635--644}.
\bibitem[{Karkus et~al.(2018)Karkus, Hsu and Lee}]{karkus2018particle}
\bibinfo{author}{Karkus, P.}, \bibinfo{author}{Hsu, D.}, \bibinfo{author}{Lee,
  W.S.}, \bibinfo{year}{2018}.
\newblock \bibinfo{title}{Particle filter networks: End-to-end probabilistic
  localization from visual observations}.
\newblock \bibinfo{journal}{arXiv preprint arXiv:1805.08975} .
\bibitem[{Karpathy et~al.(2015)Karpathy, Johnson and
  Fei-Fei}]{karpathy2015visualizing}
\bibinfo{author}{Karpathy, A.}, \bibinfo{author}{Johnson, J.},
  \bibinfo{author}{Fei-Fei, L.}, \bibinfo{year}{2015}.
\newblock \bibinfo{title}{Visualizing and understanding recurrent networks}.
\newblock \bibinfo{journal}{arXiv preprint arXiv:1506.02078} .
\bibitem[{Koul et~al.(2018)Koul, Greydanus and Fern}]{koul2018learning}
\bibinfo{author}{Koul, A.}, \bibinfo{author}{Greydanus, S.},
  \bibinfo{author}{Fern, A.}, \bibinfo{year}{2018}.
\newblock \bibinfo{title}{Learning finite state representations of recurrent
  policy networks}.
\newblock \bibinfo{journal}{arXiv preprint arXiv:1811.12530} .
\bibitem[{Li et~al.(2014)Li, Yang, Chen and Chen}]{li2014disturbance}
\bibinfo{author}{Li, S.}, \bibinfo{author}{Yang, J.}, \bibinfo{author}{Chen,
  W.H.}, \bibinfo{author}{Chen, X.}, \bibinfo{year}{2014}.
\newblock \bibinfo{title}{Disturbance observer-based control: methods and
  applications}.
\newblock \bibinfo{publisher}{CRC press}.
\bibitem[{Lillicrap et~al.(2015)Lillicrap, Hunt, Pritzel, Heess, Erez, Tassa,
  Silver and Wierstra}]{lillicrap2015continuous}
\bibinfo{author}{Lillicrap, T.P.}, \bibinfo{author}{Hunt, J.J.},
  \bibinfo{author}{Pritzel, A.}, \bibinfo{author}{Heess, N.},
  \bibinfo{author}{Erez, T.}, \bibinfo{author}{Tassa, Y.},
  \bibinfo{author}{Silver, D.}, \bibinfo{author}{Wierstra, D.},
  \bibinfo{year}{2015}.
\newblock \bibinfo{title}{Continuous control with deep reinforcement learning}.
\newblock \bibinfo{journal}{arXiv preprint arXiv:1509.02971} .
\bibitem[{Lu and Liu(2017)}]{lu2017active}
\bibinfo{author}{Lu, W.}, \bibinfo{author}{Liu, D.}, \bibinfo{year}{2017}.
\newblock \bibinfo{title}{Active task design in adaptive control of redundant
  robotic systems}, in: \bibinfo{booktitle}{Australasian Conference on Robotics
  and Automation}, \bibinfo{organization}{ARAA}.
\bibitem[{Lu and Liu(2018)}]{lu2018frequency}
\bibinfo{author}{Lu, W.}, \bibinfo{author}{Liu, D.}, \bibinfo{year}{2018}.
\newblock \bibinfo{title}{A frequency-limited adaptive controller for
  underwater vehicle-manipulator systems under large wave disturbances}, in:
  \bibinfo{booktitle}{The World Congress on Intelligent Control and
  Automation}.
\bibitem[{Lu et~al.(2015)Lu, Zhu and Ferrari}]{lu2015hybrid}
\bibinfo{author}{Lu, W.}, \bibinfo{author}{Zhu, P.}, \bibinfo{author}{Ferrari,
  S.}, \bibinfo{year}{2015}.
\newblock \bibinfo{title}{A hybrid-adaptive dynamic programming approach for
  the model-free control of nonlinear switched systems}.
\newblock \bibinfo{journal}{IEEE Transactions on Automatic Control}
  \bibinfo{volume}{61}, \bibinfo{pages}{3203--3208}.
\bibitem[{Lu et~al.(2016)Lu, Zhu and Ferrari}]{LuADPSwitched14}
\bibinfo{author}{Lu, W.}, \bibinfo{author}{Zhu, P.}, \bibinfo{author}{Ferrari,
  S.}, \bibinfo{year}{2016}.
\newblock \bibinfo{title}{An approximate dynamic programming approach for
  model-free control of switched systems}.
\newblock \bibinfo{journal}{IEEE Transactions on Automatic Control} .
\bibitem[{Mnih et~al.(2016)Mnih, Badia, Mirza, Graves, Lillicrap, Harley,
  Silver and Kavukcuoglu}]{mnih2016asynchronous}
\bibinfo{author}{Mnih, V.}, \bibinfo{author}{Badia, A.P.},
  \bibinfo{author}{Mirza, M.}, \bibinfo{author}{Graves, A.},
  \bibinfo{author}{Lillicrap, T.}, \bibinfo{author}{Harley, T.},
  \bibinfo{author}{Silver, D.}, \bibinfo{author}{Kavukcuoglu, K.},
  \bibinfo{year}{2016}.
\newblock \bibinfo{title}{Asynchronous methods for deep reinforcement
  learning}, in: \bibinfo{booktitle}{International conference on machine
  learning}, pp. \bibinfo{pages}{1928--1937}.
\bibitem[{Mnih et~al.(2015)Mnih, Kavukcuoglu, Silver, Rusu, Veness, Bellemare,
  Graves, Riedmiller, Fidjeland, Ostrovski et~al.}]{mnih2015human}
\bibinfo{author}{Mnih, V.}, \bibinfo{author}{Kavukcuoglu, K.},
  \bibinfo{author}{Silver, D.}, \bibinfo{author}{Rusu, A.A.},
  \bibinfo{author}{Veness, J.}, \bibinfo{author}{Bellemare, M.G.},
  \bibinfo{author}{Graves, A.}, \bibinfo{author}{Riedmiller, M.},
  \bibinfo{author}{Fidjeland, A.K.}, \bibinfo{author}{Ostrovski, G.}, et~al.,
  \bibinfo{year}{2015}.
\newblock \bibinfo{title}{Human-level control through deep reinforcement
  learning}.
\newblock \bibinfo{journal}{Nature} \bibinfo{volume}{518},
  \bibinfo{pages}{529}.
\bibitem[{Nagabandi et~al.(2018)Nagabandi, Kahn, Fearing and
  Levine}]{nagabandi2018neural}
\bibinfo{author}{Nagabandi, A.}, \bibinfo{author}{Kahn, G.},
  \bibinfo{author}{Fearing, R.S.}, \bibinfo{author}{Levine, S.},
  \bibinfo{year}{2018}.
\newblock \bibinfo{title}{Neural network dynamics for model-based deep
  reinforcement learning with model-free fine-tuning}, in:
  \bibinfo{booktitle}{Robotics and Automation (ICRA), 2018 IEEE International
  Conference on}, \bibinfo{organization}{IEEE}. pp.
  \bibinfo{pages}{7579--7586}.
\bibitem[{Oh et~al.(2016)Oh, Chockalingam, Singh and Lee}]{oh2016control}
\bibinfo{author}{Oh, J.}, \bibinfo{author}{Chockalingam, V.},
  \bibinfo{author}{Singh, S.}, \bibinfo{author}{Lee, H.}, \bibinfo{year}{2016}.
\newblock \bibinfo{title}{Control of memory, active perception, and action in
  minecraft}.
\newblock \bibinfo{journal}{arXiv preprint arXiv:1605.09128} .
\bibitem[{Ohishi et~al.(1987)Ohishi, Nakao, Ohnishi and
  Miyachi}]{ohishi1987microprocessor}
\bibinfo{author}{Ohishi, K.}, \bibinfo{author}{Nakao, M.},
  \bibinfo{author}{Ohnishi, K.}, \bibinfo{author}{Miyachi, K.},
  \bibinfo{year}{1987}.
\newblock \bibinfo{title}{Microprocessor-controlled dc motor for
  load-insensitive position servo system}.
\newblock \bibinfo{journal}{IEEE Transactions on Industrial Electronics} ,
  \bibinfo{pages}{44--49}.
\bibitem[{Omlin and Giles(1992)}]{omlin1992training}
\bibinfo{author}{Omlin, C.W.}, \bibinfo{author}{Giles, C.L.},
  \bibinfo{year}{1992}.
\newblock \bibinfo{title}{Training second-order recurrent neural networks using
  hints}, in: \bibinfo{booktitle}{Machine Learning Proceedings 1992}.
  \bibinfo{publisher}{Elsevier}, pp. \bibinfo{pages}{361--366}.
\bibitem[{Paull and Unger(1959)}]{paull1959minimizing}
\bibinfo{author}{Paull, M.C.}, \bibinfo{author}{Unger, S.H.},
  \bibinfo{year}{1959}.
\newblock \bibinfo{title}{Minimizing the number of states in incompletely
  specified sequential switching functions}.
\newblock \bibinfo{journal}{IRE Transactions on Electronic Computers} ,
  \bibinfo{pages}{356--367}.
\bibitem[{Ranganathan and Campbell(2003)}]{ranganathan2003middleware}
\bibinfo{author}{Ranganathan, A.}, \bibinfo{author}{Campbell, R.H.},
  \bibinfo{year}{2003}.
\newblock \bibinfo{title}{A middleware for context-aware agents in ubiquitous
  computing environments}, in: \bibinfo{booktitle}{ACM/IFIP/USENIX
  International Conference on Distributed Systems Platforms and Open
  Distributed Processing}, \bibinfo{organization}{Springer}. pp.
  \bibinfo{pages}{143--161}.
\bibitem[{Read(2011)}]{read2011bp}
\bibinfo{author}{Read, C.}, \bibinfo{year}{2011}.
\newblock \bibinfo{title}{BP and the Macondo spill: the complete story}.
\newblock \bibinfo{publisher}{Springer}.
\bibitem[{S{\ae}mundsson et~al.(2018)S{\ae}mundsson, Hofmann and
  Deisenroth}]{saemundsson2018meta}
\bibinfo{author}{S{\ae}mundsson, S.}, \bibinfo{author}{Hofmann, K.},
  \bibinfo{author}{Deisenroth, M.P.}, \bibinfo{year}{2018}.
\newblock \bibinfo{title}{Meta reinforcement learning with latent variable
  gaussian processes}.
\newblock \bibinfo{journal}{arXiv preprint arXiv:1803.07551} .
\bibitem[{Samek et~al.(2017)Samek, Wiegand and
  M{\"u}ller}]{samek2017explainable}
\bibinfo{author}{Samek, W.}, \bibinfo{author}{Wiegand, T.},
  \bibinfo{author}{M{\"u}ller, K.R.}, \bibinfo{year}{2017}.
\newblock \bibinfo{title}{Explainable artificial intelligence: Understanding,
  visualizing and interpreting deep learning models}.
\newblock \bibinfo{journal}{arXiv preprint arXiv:1708.08296} .
\bibitem[{Schulman et~al.(2015a)Schulman, Levine, Abbeel, Jordan and
  Moritz}]{schulman2015trust}
\bibinfo{author}{Schulman, J.}, \bibinfo{author}{Levine, S.},
  \bibinfo{author}{Abbeel, P.}, \bibinfo{author}{Jordan, M.},
  \bibinfo{author}{Moritz, P.}, \bibinfo{year}{2015}a.
\newblock \bibinfo{title}{Trust region policy optimization}, in:
  \bibinfo{booktitle}{International Conference on Machine Learning}, pp.
  \bibinfo{pages}{1889--1897}.
\bibitem[{Schulman et~al.(2015b)Schulman, Moritz, Levine, Jordan and
  Abbeel}]{schulman2015high}
\bibinfo{author}{Schulman, J.}, \bibinfo{author}{Moritz, P.},
  \bibinfo{author}{Levine, S.}, \bibinfo{author}{Jordan, M.},
  \bibinfo{author}{Abbeel, P.}, \bibinfo{year}{2015}b.
\newblock \bibinfo{title}{High-dimensional continuous control using generalized
  advantage estimation}.
\newblock \bibinfo{journal}{arXiv preprint arXiv:1506.02438} .
\bibitem[{Skogestad and Postlethwaite(2007)}]{skogestad2007multivariable}
\bibinfo{author}{Skogestad, S.}, \bibinfo{author}{Postlethwaite, I.},
  \bibinfo{year}{2007}.
\newblock \bibinfo{title}{Multivariable feedback control: analysis and design}.
  volume~\bibinfo{volume}{2}.
\newblock \bibinfo{publisher}{Wiley New York}.
\bibitem[{Umeno et~al.(1993)Umeno, Kaneko and Hori}]{umeno1993robust}
\bibinfo{author}{Umeno, T.}, \bibinfo{author}{Kaneko, T.},
  \bibinfo{author}{Hori, Y.}, \bibinfo{year}{1993}.
\newblock \bibinfo{title}{Robust servosystem design with two degrees of freedom
  and its application to novel motion control of robot manipulators}.
\newblock \bibinfo{journal}{IEEE Transactions on Industrial Electronics}
  \bibinfo{volume}{40}, \bibinfo{pages}{473--485}.
\bibitem[{Wang et~al.(2019)Wang, Lu, Yan and Liu}]{wang2019dob}
\bibinfo{author}{Wang, T.}, \bibinfo{author}{Lu, W.}, \bibinfo{author}{Yan,
  Z.}, \bibinfo{author}{Liu, D.}, \bibinfo{year}{2019}.
\newblock \bibinfo{title}{Dob-net: Actively rejecting unknown excessive
  time-varying disturbances}.
\newblock \bibinfo{journal}{arXiv preprint arXiv:1907.04514} .
\bibitem[{Waslander and Wang(2009)}]{waslander2009wind}
\bibinfo{author}{Waslander, S.}, \bibinfo{author}{Wang, C.},
  \bibinfo{year}{2009}.
\newblock \bibinfo{title}{Wind disturbance estimation and rejection for
  quadrotor position control}, in: \bibinfo{booktitle}{AIAA Infotech@ Aerospace
  Conference and AIAA Unmanned... Unlimited Conference}, p.
  \bibinfo{pages}{1983}.
\bibitem[{Weiss et~al.(2017)Weiss, Goldberg and Yahav}]{weiss2017extracting}
\bibinfo{author}{Weiss, G.}, \bibinfo{author}{Goldberg, Y.},
  \bibinfo{author}{Yahav, E.}, \bibinfo{year}{2017}.
\newblock \bibinfo{title}{Extracting automata from recurrent neural networks
  using queries and counterexamples}.
\newblock \bibinfo{journal}{arXiv preprint arXiv:1711.09576} .
\bibitem[{Woolfrey et~al.(2016)Woolfrey, Liu and
  Carmichael}]{woolfrey2016kinematic}
\bibinfo{author}{Woolfrey, J.}, \bibinfo{author}{Liu, D.},
  \bibinfo{author}{Carmichael, M.}, \bibinfo{year}{2016}.
\newblock \bibinfo{title}{Kinematic control of an autonomous underwater
  vehicle-manipulator system (auvms) using autoregressive prediction of vehicle
  motion and model predictive control}, in: \bibinfo{booktitle}{Robotics and
  Automation (ICRA), 2016 IEEE International Conference on},
  \bibinfo{organization}{IEEE}. pp. \bibinfo{pages}{4591--4596}.
\bibitem[{Woolfrey et~al.(2019)Woolfrey, Lu and Liu}]{woolfrey2019control}
\bibinfo{author}{Woolfrey, J.}, \bibinfo{author}{Lu, W.}, \bibinfo{author}{Liu,
  D.}, \bibinfo{year}{2019}.
\newblock \bibinfo{title}{A control method for joint torque minimization of
  redundant manipulators handling large external forces}.
\newblock \bibinfo{journal}{Journal of Intelligent \& Robotic Systems} ,
  \bibinfo{pages}{1--14}.
\bibitem[{Xie and Guo(2000)}]{xie2000much}
\bibinfo{author}{Xie, L.L.}, \bibinfo{author}{Guo, L.}, \bibinfo{year}{2000}.
\newblock \bibinfo{title}{How much uncertainty can be dealt with by feedback?}
\newblock \bibinfo{journal}{IEEE Transactions on Automatic Control}
  \bibinfo{volume}{45}, \bibinfo{pages}{2203--2217}.
\bibitem[{Yang et~al.(2010)Yang, Li, Chen and Li}]{yang2010disturbance}
\bibinfo{author}{Yang, J.}, \bibinfo{author}{Li, S.}, \bibinfo{author}{Chen,
  X.}, \bibinfo{author}{Li, Q.}, \bibinfo{year}{2010}.
\newblock \bibinfo{title}{Disturbance rejection of ball mill grinding circuits
  using dob and mpc}.
\newblock \bibinfo{journal}{Powder Technology} \bibinfo{volume}{198},
  \bibinfo{pages}{219--228}.
\bibitem[{Yuan and Wu(2015)}]{yuan2015switching}
\bibinfo{author}{Yuan, C.}, \bibinfo{author}{Wu, F.}, \bibinfo{year}{2015}.
\newblock \bibinfo{title}{Switching control of linear systems subject to
  asymmetric actuator saturation}.
\newblock \bibinfo{journal}{International Journal of Control}
  \bibinfo{volume}{88}, \bibinfo{pages}{204--215}.
\bibitem[{Zoph and Le(2016)}]{zoph2016neural}
\bibinfo{author}{Zoph, B.}, \bibinfo{author}{Le, Q.V.}, \bibinfo{year}{2016}.
\newblock \bibinfo{title}{Neural architecture search with reinforcement
  learning}.
\newblock \bibinfo{journal}{arXiv preprint arXiv:1611.01578} .
\bibitem[{Zuo et~al.(2010)Zuo, Ho and Wang}]{zuo2010fault}
\bibinfo{author}{Zuo, Z.}, \bibinfo{author}{Ho, D.W.}, \bibinfo{author}{Wang,
  Y.}, \bibinfo{year}{2010}.
\newblock \bibinfo{title}{Fault tolerant control for singular systems with
  actuator saturation and nonlinear perturbation}.
\newblock \bibinfo{journal}{Automatica} \bibinfo{volume}{46},
  \bibinfo{pages}{569--576}.

\end{thebibliography}


%
%

\end{document}